\documentclass[letterpaper]{article} % DO NOT CHANGE THIS
\usepackage{aaai25}  % DO NOT CHANGE THIS
\usepackage{times}  % DO NOT CHANGE THIS
\usepackage{helvet}  % DO NOT CHANGE THIS
\usepackage{courier}  % DO NOT CHANGE THIS
\usepackage[hyphens]{url}  % DO NOT CHANGE THIS
\usepackage{graphicx} % DO NOT CHANGE THIS
\usepackage{natbib}  % DO NOT CHANGE THIS AND DO NOT ADD ANY OPTIONS TO IT
\usepackage{caption} % DO NOT CHANGE THIS AND DO NOT ADD ANY OPTIONS TO IT
\usepackage{algorithmic}
\usepackage[ruled,vlined,linesnumbered]{algorithm2e}
\usepackage{booktabs}
\usepackage{xcolor}
\usepackage{amsmath}
\usepackage{amssymb}

\newcommand{\frameworkname}[0]{TF-GPH}
\newcommand{\SubfigColor}{red}
\newcommand{\Figref}[1]{Fig.\textcolor{red}{~\ref{#1}}}
\newcommand{\Tabref}[1]{Tab.\textcolor{red}{~\ref{#1}}}
\newcommand{\Eqref}[1]{Eq.\textcolor{red}{~\eqref{#1}}}
\newcommand{\Algref}[1]{Alg.\textcolor{red}{~\ref{#1}}}

\definecolor{comment_green}{rgb}{0.17254, 0.47058, 0.39607}

\usepackage{newfloat}
\usepackage{listings}
\DeclareCaptionStyle{ruled}{labelfont=normalfont,labelsep=colon,strut=off} % DO NOT CHANGE THIS
\title{Training-and-Prompt-Free General Painterly Harmonization via Zero-Shot Disentenglement on Style and Content References}
\author{
    Teng-Fang Hsiao,
    Bo-Kai Ruan,
    Hong-Han Shuai
}
\affiliations{
    National Yang Ming Chiao Tung University, Taiwan\\
    \{tfhsiao.ee13, bkruan.ee11, hhshuai\}@nycu.edu.tw
}

\usepackage{bibentry}
\begin{document}

\maketitle
\begin{abstract}
Painterly image harmonization aims at seamlessly blending disparate visual elements within a single image. However, previous approaches often struggle due to limitations in training data or reliance on additional prompts, leading to inharmonious and content-disrupted output. To surmount these hurdles, we design a Training-and-prompt-Free General Painterly Harmonization method (TF-GPH). TF-GPH incorporates a novel ``Similarity Disentangle Mask'', which disentangles the foreground content and background image by redirecting their attention to corresponding reference images, enhancing the attention mechanism for multi-image inputs. Additionally, we propose a ``Similarity Reweighting'' mechanism to balance harmonization between stylization and content preservation. This mechanism minimizes content disruption by prioritizing the content-similar features within the given background style reference. Finally, we address the deficiencies in existing benchmarks by proposing novel range-based evaluation metrics and a new benchmark to better reflect real-world applications. Extensive experiments demonstrate the efficacy of our method in all benchmarks. More detailed in \textcolor{blue}{\url{https://github.com/BlueDyee/TF-GPH}}.
\end{abstract}

\section{Introduction}

Image composition, which involves blending a foreground element from one image with a different background, often results in composite images with mismatched colors and illumination between the foreground and background. Image harmonization techniques have been developed to adjust the appearance of foreground for a seamless integration with the background \cite{DIH,GP_GAN,DIM_dual,composite_photograph}. A specialized area within this field, painterly image harmonization, focuses on integrating elements into paintings to enable artistic edits \cite{PHDiff,DPH}. For instance, ProPIH \cite{propih}, pioneers progressive painterly harmonization, training the model with different levels of harmonization, enhancing its applicability to real-world scenarios.

\begin{figure}[h]
  \centering
  \center
  \includegraphics[width=0.4\textwidth]{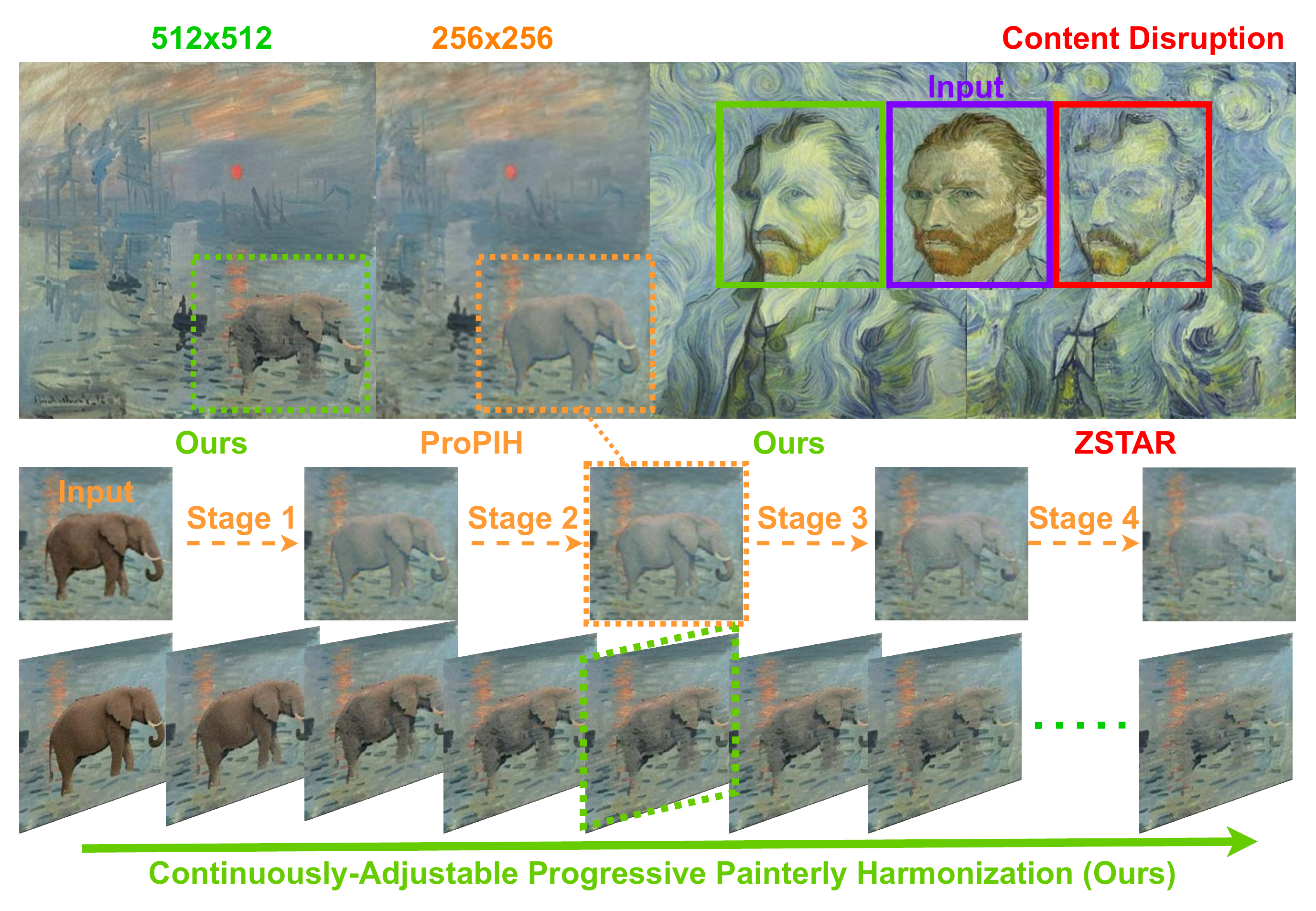}
  \caption{Our method overcomes the resolution and staged-progressive painterly harmonization limitations present in the SOTA method ProPIH \cite{propih}, where users are restricted to selecting stylization strength from one of four stages. In contrast, our approach offer continuously adjustable hyperparameters, allowing for more flexible stylization.  Additionally, our method effectively mitigates content disruption issues, such as facial alterations, commonly seen in image-editing methods like ZSTAR \cite{zstar}.}
  \label{fig:small_teaser}
\end{figure}

Despite notable advancements, current painterly image harmonization techniques still face challenges with generalizability, particularly when dealing with novel art styles or unique content compositions. One promising solution is to leverage insights from other image-editing methods. For instance, \cite{InST,VCT} suggest fine-tuning models to adapt to input styles. However, each styles require additional computational costs that are 10x times higher than a single inference. Alternatively, \cite{tficon, clipstyler} propose text-guided editing strategies, but these approaches are limited by the difficulty of adequately describing complex visual styles through text alone. Recently, methods such as \cite{masactrl,  zstar} explore training-free techniques. These methods utilize attention-sharing across images combined with techniques like AdaIN\cite{adaIN} to align content features with style references. While effective, this brute alignment lead to content disruption as shown in \Figref{fig:small_teaser}.

  \begin{figure*}[h]
  \centering
  \center
  \includegraphics[width=0.95\textwidth]{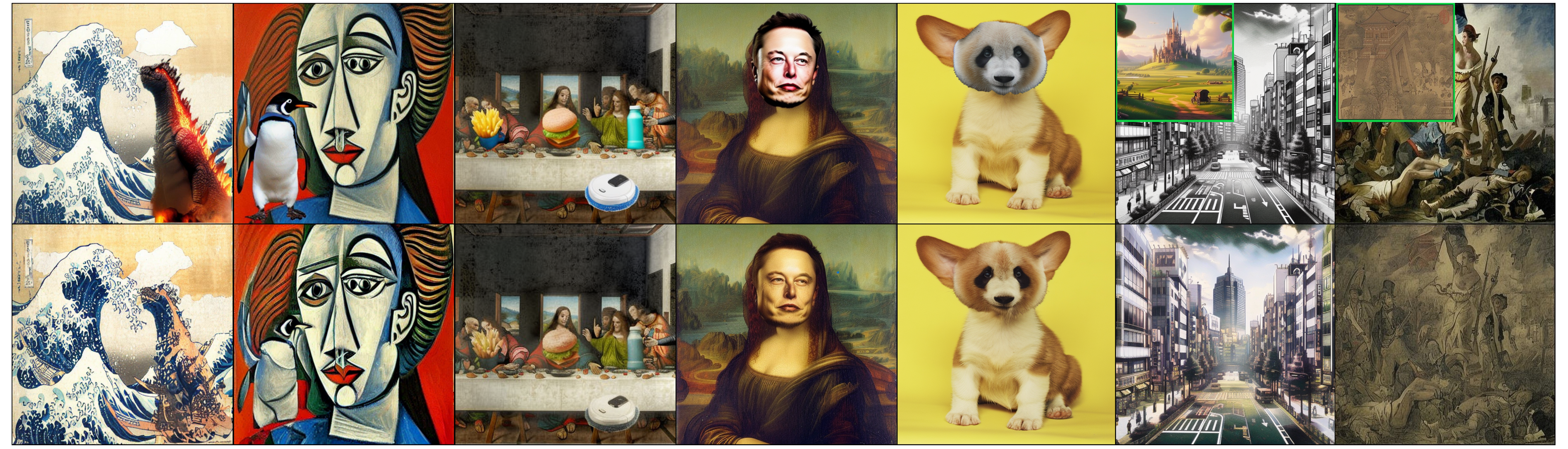}
  \caption{An example demonstrates three tasks in general painterly harmonization: Object Insertion (columns 1 to 3), Object Swapping (columns 4 and 5), and Style Transfer (columns 6 and 7). The top row features user-generated composite images, where \textcolor{green}{green boxes} highlighting the style reference of final two. The bottom row showcases the results using our method.}
  \label{fig:teaser}
\end{figure*}

% v1
In this work, we present \textbf{\frameworkname{}}, an innovative diffusion pipeline that operates without additional training or prompts by leveraging the pretrained diffusion model \cite{LDM}. \frameworkname{} solves a wider range of painterly harmonization tasks, including object insertion, swapping, and style transfers, as illustrated in \Figref{fig:teaser}. In our pipeline, we modify the self-attention mechanism of the diffusion model and adopt the shared attention layer from \cite{style_aligned} to enable multi-image input (foreground content reference, background style reference, and composite image). While the shared attention layer can merge similar features across images, such as consecutive frames, the strong self-similarity of the composite image causes it to overlook dissimilar references in this task. To address this, we introduce a novel \textbf{similarity disentangle mask} within the shared attention layer. This mask applied before the softmax operation decouples the foreground and background features by redirecting the composite image's self-attention to the two reference images. This approach allows precise control over the foreground and background within the composite image by adjusting the attention to their respective references.

Moreover, to address the content disruptions caused by attention adjustments such as addition and AdaIN, as observed in prior work, we propose a similarity-based editing method termed \textbf{similarity reweighting}. This approach balances attention between content and style references by scaling similarity based on user specified hyperparameters. By prioritizing style features that closely match the content features the content disruption thus minimized. By integrating these two aforementioned adjustments into the existing image generation pipeline, we are able to perform image harmonization without requiring additional prompts or training. Additionally, this mechanism offering flexibility to tailor the output continuously from style-free to style-heavy, thereby accommodating various artistic preferences, as shown in \Figref{fig:small_teaser}.

Finally, a challenge in evaluating painterly harmonization is the limited diversity of test data styles \cite{wikiart}, which are often restricted to those seen during training. This limitation fails to capture the wide variety of styles encountered in real-world scenarios, such as manga or cartoon. To address this issue, we introduce the ``General Painterly Harmonization Benchmark'' (GPH Benchmark). This benchmark encompasses three harmonization tasks—object insertion, object swapping, and style transfer—while incorporating diverse content and style references to ensure a comprehensive evaluation. Furthermore, existing metrics typically focus on either content or style similarity without adequately reflecting user preferences for different balances between stylization and content preservation. To bridge this gap, we propose range-based metrics, evaluating both the lower and upper bounds of stylization and content-preservation strength across the dataset. A wider range indicates greater flexibility and the adaptability to various scenarios. 

Our contributions can be summarized as follows. 1) We introduce the \textbf{\frameworkname{}} framework, the first training-and-prompt-free pipeline using a diffusion model designed for general painterly harmonization. 2) Our proposed \textbf{similarity disentangle mask with similarity reweighting} not only shows promising results in painterly harmonization tasks, but also solves the content disruption issue of existing attention-based editing method. 3)We propose the \textbf{GPH Benchmark}, consisting of various data for real-world usage, together with a range-based metric to align model performance with user experience.

\section{Related Work}
  \subsection{Image Harmonization}Image Harmonization can be categorized into two main types: \textbf{Realistic Harmonization} and \textbf{Painterly Harmonization}. The former \cite{DIC,dovenet,ST_GAN,hierachical_harmonization} focus on seamlessly integrating objects into new backgrounds with consistent illumination, edge alignment, and shadow integrity. In contrast, Painterly Harmonization \cite{PHDiff,PHDNet,GP_GAN} aims to artistically blend objects into paintings, prioritizing stylistic coherence. Recently, ArtoPIH \cite{artopih} propose learning from painterly objects by using annotated objects within paintings as training data. Additionally, ProPIH \cite{propih} introduce a progressive learning approach, improving practical applicability. Despite these advancements, existing methods require training, which can limit their usability. Our proposed method, however, eliminates the need for training, enabling direct application to unseen styles and significantly enhancing the versatility of painterly image harmonization.

  \subsection{Attention-based Image Editing} Manipulation of attention layers within diffusion UNet architectures is a prevalent strategy in modern image editing techniques \cite{plug_and_play,attention_excite,photoswap,tficon,P2P}. For instance, P2P \cite{P2P} utilizes prompt-driven cross-attention to modify images, while TF-ICON \cite{tficon} integrates objects into backgrounds by constraining self-attention and cross-attention outputs with given mask. Despite their effectiveness, the reliance on descriptive prompts can be problematic when suitable prompts are not available. In contrast, our \frameworkname{} method function solely with image inputs, eliminating the need of prompts.
  
  \subsection{Style Transfer} Style transfer aims to alter the style of a content image to match a specified style. Existing methods generally categorized into optimization-based and feedforward-based approaches. The former \cite{gatys,demystifying}, refine the image by aligning it with features extracted from the style reference. For example, \cite{clipstyler} utilizes a pre-trained CLIP model \cite{CLIP} for this purpose. In contrast, feedforward-based \cite{StyTr2,QuantArt} involving VCT \cite{VCT} and InST \cite{InST}, which fine-tune models to integrate style into the model's architecture. Recently, attention-based techniques have been incorporated. For instance, the shared attention module introduced in \cite{style_aligned, zstar,style_injection} produces feature-consistent images by sharing attention across multiple images. However, these methods often suffer from content disruption due to the blending of unrelated features from different references. In contrast, \frameworkname{} minimizes content disruption and achieves superior performance across styles.

\begin{figure*}[h]
  \centering
  \includegraphics[width=0.91\linewidth]{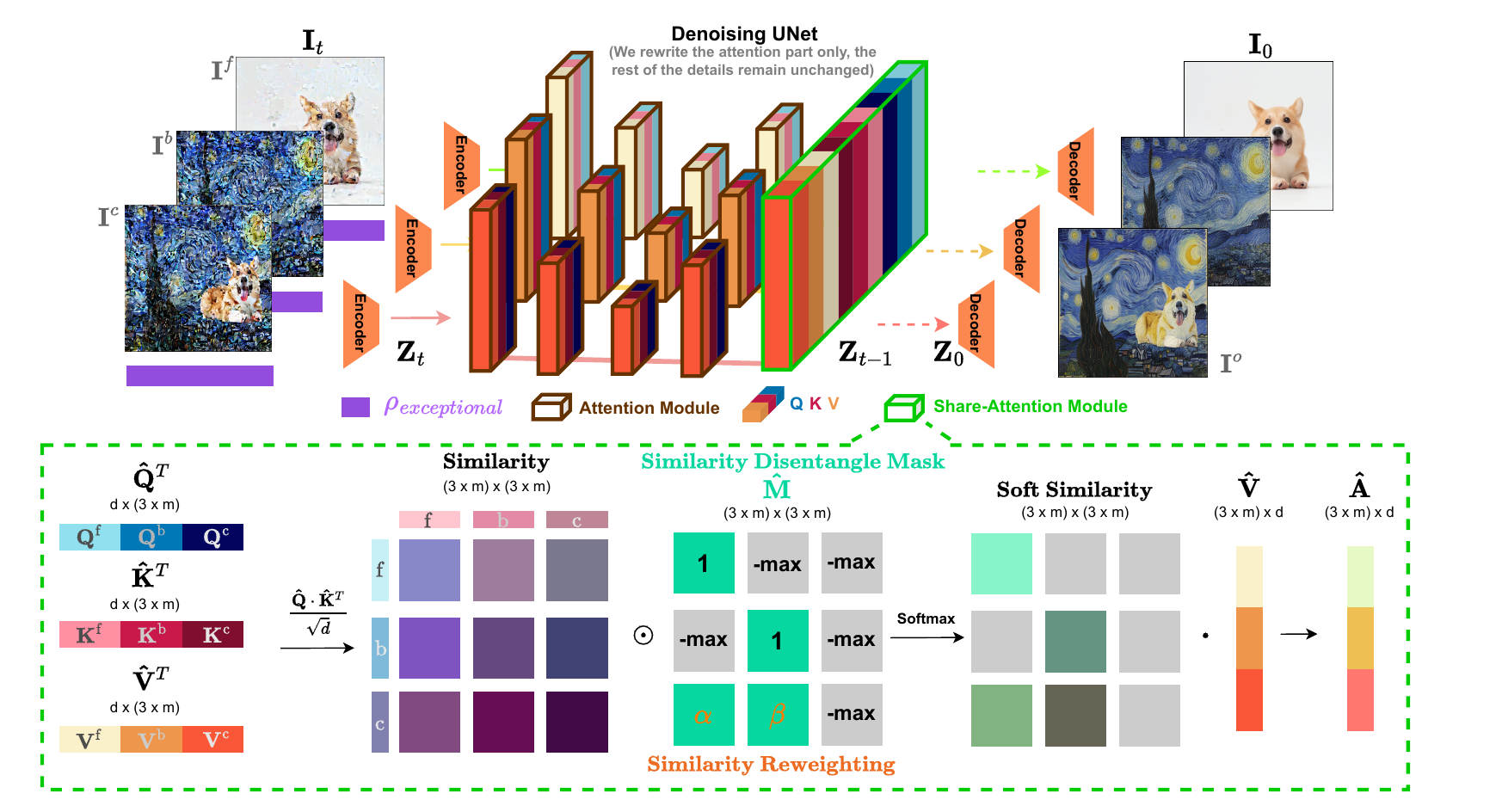}
  \caption{The architecture of our proposed \frameworkname{} method involves several stages. Initially, we feed the denoising U-Net with the inverse latent $Z_t$, and during the first $l < L_{\text{share}}-1$ layers of the U-Net, the three latent representations, $z^\text{f}_t$, $z^\text{b}_t$, and $z^\text{c}_t$, are forwarded separately to the Attention Module. Afterward, they are fed into the \textcolor{blue}{Share-Attention Module} (the \textcolor{blue}{blue} part below), obtaining their image-wise attention via \Eqref{eq:share_attention}. In the end, the output harmonized image $I^\text{o}$ is produced.}
  \label{fig:overall}
\end{figure*}

\section{Method}
    Our research aims to facilitate a general form of painterly harmonization based on images only without the additional need for prompts, which can facilitate various applications, \textit{i.e.}, object insertion, object swapping, and style transfer. Formally, given a foreground object image $I^\text{f}$, a background painting $I^\text{b}$, and $I^\text{c}$, which is the user-specified composition that guides the size and position of the foreground object on the background painting, the goal of painterly harmonization is to transfer the style from $I^\text{b}$ to the object from $I^\text{f}$ in $I^\text{c}$ seamlessly, resulting a harmonized image $I^\text{o}$.

    To address the challenge of painterly harmonization, we introduce a novel framework titled Training-and-Prompt-Free General Painterly Harmonization (\frameworkname{}), as depicted in \Figref{fig:overall}. Specifically, the inputs—foreground $I^\text{f}$, background $I^\text{b}$, and composite $I^{\text{c}}$-are initially processed through an inversion mechanism equipped with either a null prompt embedding or a exceptional prompt embedding $\rho_{\text{exceptional}}$, which has demonstrated its ability for stabilizing inversion process \cite{tficon}. Subsequently, a denoising operation is applied concurrently to all three images, during which the composite image $I^{\text{c}}$ is enriched with style attributes, producing harmonized output $I^{\text{o}}$

    %Modified
    %\textcolor{red}{
    The core of our architecture is the \textbf{Similarity Disentangle Mask}, a novel attention mask designed to disentangle the features of the foreground object from the background image of $I^\text{c}$ and link them to their corresponding references $I^\text{f}$ and $I^\text{b}$. After disentanglement, we enhance the influence of the background style reference $I^\text{b}$ on the pasted object through our \textbf{Similarity Reweighting} technique. This approach differs from existing attention-based editing techniques, which directly add \cite{P2P} or adjust the mean/variance of features \cite{style_aligned, style_injection}, introducing disruption on semantic and structural details. By adjusting the similarity solely, we can minimize content disruption while applying the stylization effect, producing the final painterly harmonized output image $I^\text{o}$. Additionally, our framework is versatile enough to support not only painterly harmonization for object insertion—a traditional task of painterly harmonization—but also object swapping and style transfer. The former is viewed as a semantically richer variant of object insertion, and the latter as a broader aspect of the same. We summarize these related tasks under the term ``General Painterly Harmonization''.
    %}
    %Note that the superscript $"in"$ disappears because the reconstructed and harmonized image $x^{\text{c}}$ is different from the input image $x^{\text{comp,in}}$. 

    % Our methodology diverges from the prior art, which requires masking \cite{PHD,PHDnet,DIB} or prompting information \cite{tficon, SDedit, VCT}.
    %To the best of our knowledge, our method is the first image stylization approach capable of cold start.
    
  \subsection{Attention-based Diffusion UNet}  
    In the framework of diffusion models, the attention mechanisms~\cite{attention} are essential to capture characteristic details, facilitating both the elimination of noise and the enhancement of context information. Specifically, the self-attention module plays a vital role in synthesizing the output by internalizing the inherent data characteristics, while the cross-attention module is instrumental in incorporating contextual information from various modalities, \textit{e.g.} text and audio, thus amplifying the conditional impact on the resultant images. Since our approach does not need an additional prompt to guide the fusion, we can simply utilize self-attention to ensure that the background style is harmonically fused into the composition image during denoising.

    Given three input images $I^\text{f}$, $I^\text{b}$, and $I^\text{c}$, these images are first compressed by a VAE encoder~\cite{LDM} into latent representations $z^\text{f}_0\in \mathbb{R}^{w \times h \times  d}$, $z^\text{b}_0\in \mathbb{R}^{w \times h \times  d}$, $z^\text{c}_0\in \mathbb{R}^{w \times h \times  d}$, respectively, where $w$ and $h$ denote the width and height of the latent shape, $d$ is the feature channels and the subscript $0$ denotes the initial timestep of the diffusion process. Next, we apply the DPM-Solver++ inversion to convert the initial latents $z^\text{f}_0$, $z^\text{b}_0$, and $z^\text{c}_0$ to noisy latents $z^\text{f}_T$, $z^\text{b}_T$, and $z^\text{c}_T$. This preprocess enabling the image modification during subsequent reconstruction process.
    
    %To related
    % Share attention module (named by \cite{style_aligned}), is a newly introduced practice for producing features consistent images by sharing the attention QKV with other images, this methods is widely adopted in the realm of video generation \cite{tokenflow, tune_a_video} and image synthesis \cite{zstar,style_injection}
    \subsection{Share-Attention Module}
    During the reconstruction process from the time step $T$ to $0$, we incorporate the style feature into $z^{\text{c}}_t$ using the shared attention module. This module can be viewed as a more general form of the self-attention module, allowing for feature flow between input images. Specifically, the traditional self-attention module projects the input feature $z\in \mathbb{R}^ {m \times  d}$ of length $m = (w \cdot h)$ onto the corresponding $Q,K,V \in \mathbb{R}^{m \times d}$ through learned linear layers inside the original self-attention module and computes the attention matrix $A \in \mathbb{R}^{m \times d}$ as follows.
    \begin{equation}
    A(Q, K, V) = \text{Softmax}\left(QK^T/\sqrt{d}\right)V,
    \label{eq:basic_attention}
    \end{equation}
    
    To enable the flow of feature information between images during the attention operation, we should consider three images at the same time instead of processing each attention matrix independently. To create the query, key, and value from three different inputs, we first concatenate three input latents on the first dimension to form $Z_T\in \mathbb{R}^{(3 \cdot m) \times d}$ = [$z^\text{f}_T$, $z^\text{b}_T$, $z^\text{c}_T$]. Then we project the latent $Z_T$ into the corresponding $\hat{Q},\hat{K},\hat{V} \in \mathbb{R}^{(3 \cdot m) \times d}$.

    %\textcolor{red}{
    \subsection{Similarity Disentangle Mask}
    However, directly feeding $\hat{Q},\hat{K},\hat{V}$ into Eq.~\eqref{eq:basic_attention} may disrupt the features of $z^\text{f}$ and $z^\text{b}$ since the additional attention from other images making the latent differ from original reconstruction without attention from others. To keep the content of $z^\text{f}$ and $z^\text{b}$ intact for correctly guiding the harmonization of $z^{\text{c}}$, we propose a specially designed mask called \textbf{similarity disentangle mask} $\hat{M} \in \mathbb{R}^{(3\cdot m)\times (3\cdot m)}$ that allows $z^{\text{c}}$ to utilize information from $z^\text{f}$ and $z^\text{b}$ while keeping $z^{\text{f}}$ and $z^{\text{b}}$ intact. The shared attention equation is thus calculated by:
    %}
    \begin{equation}
    \hat{A}(\hat{Q}, \hat{K}, \hat{V}) = \text{Softmax}\left(\hat{M}\odot (\hat{Q}\hat{K}^T)/\sqrt{d}\right)\hat{V},
    \label{eq:share_attention}
    \end{equation}
    where $\odot$ denotes the Hadamard product. Afterward, Eq.\eqref{eq:share_attention} outputs the batch attention $\hat{A}\in \mathbb{R}^{(3\cdot m) \times d}$ containing the intact $A^{\text{f}}$, $A^{\text{b}}$, and $A^{\text{c}}$ guided by the features of $z^\text{b}$ and $z^\text{b}$.
    The specially designed $\hat{M}$ can be visualized as:
    %can be reshaped into $M \in \mathbb{R}^{3 \times 3 \times (m \times m)}$ for better understanding:
    \[
    \hat{M} = \begin{bmatrix}
    1 \cdot J & \nu \cdot J & \nu \cdot J \\
    \nu \cdot J & 1 \cdot J & \nu \cdot J \\
    \alpha \cdot J & \beta \cdot J & \gamma \cdot J \\
    \end{bmatrix}
    \]
    Here, $J \in \mathbb{R}^{m \times m}$ is an all-one matrix, and $\nu=-\infty$ minimizes the similarity between $Q$ and $K$ on the corresponding entry, keeping $A^{\text{f}}$, $A^{\text{b}}$ intact. While $\alpha$, $\beta$, and $\gamma$ control the attention of $Q^{\text{c}}$ towards $K^{\text{f}}$, $K^{\text{b}}$, and $K^{\text{c}}$, respectively.. It is worth noting that when setting $\alpha=-\infty$, $\beta=-\infty$, and $\gamma=1$, each row in Eq.\eqref{eq:share_attention} is equivalent to Eq. \eqref{eq:basic_attention} as $Q^{\text{c}}$, $Q^{\text{f}}$, and $Q^{\text{b}}$ can only attend to its counterpart $K^{\text{c}}$, $K^{\text{f}}$, and $K^{b}$ without information from other images. Therefore, our proposed similarity disentangle mask can be viewed as an expansion of attention mechanism with adjustable entries controlling features sharing.

\begin{figure}[t!]
      \centering
      \includegraphics[width=0.8\linewidth]{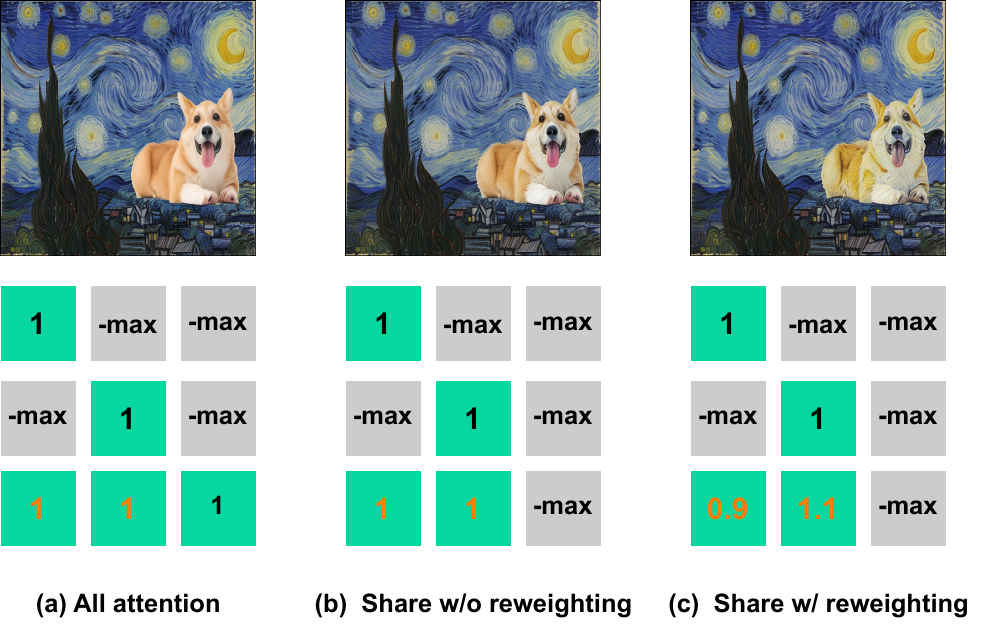}
      \caption{Comparisons of different attention strategy with corresponding similarity mask (read with \Figref{fig:overall}).}
      \label{fig:attention_strategy}
    \end{figure}

    % Modified
    %\textcolor{red}{
    Furthermore, to completely disentangle the features related to the object reference $z^\text{f}$ from $z^\text{b}$, we set the entry $\gamma$ to $-\infty$, which blocks the functionality of $K^\text{c}$ and $V^\text{c}$. By this means, we can control the features related to the pasted-foreground object within $z^\text{c}$ by modifying entry $\alpha$, which control the influence of $z^\text{f}$, and similarly, control the background features within $z^\text{c}$ related to $z^\text{b}$ by adjusting its corresponding entry $\beta$. As shown in \Figref{fig:attention_strategy}\textcolor{\SubfigColor}{(b)}, the output remains nearly the same to \Figref{fig:attention_strategy}\textcolor{\SubfigColor}{(a)} validating that the features of $z^\text{c}$ can be totally controlled by two references $z^\text{f}$ and $z^\text{b}$ 
    %}
    
    \subsection{Similarity Reweighting}\label{sec:method-reweighting}
    Another intriguing observation from \Figref{fig:attention_strategy}\textcolor{\SubfigColor}{(a) and (b)} is that the output image only changes slightly even when the pasted ``corgi'' in $I^{\text{c}}$ has a different resolution compared to the ``corgi'' in $I^{f}$. We infer that the robustness of the pretrained diffusion model enables it to capture high-level semantic and structural information despite minor disturbances, such as differences in scale and position. Consequently, the self-attention layer can withstand these perturbations, producing results that remain similar to the original input.

    % Modified
    %\textcolor{red}{
    To determine the perturbation that can break the self-attention robustness while still generating high-quality results, a simple yet effective idea of attention injection has been widely adopted by previous research \cite{photoswap, P2P, tficon, plug_and_play}. This approach introduces strong perturbations to the attention mechanism by directly appending either prompt-guided cross-attention output or image-guided self-attention output computed with other images. Another common strategy is applying the ``AdaIN'' technique to different components. For example, \cite{style_injection} compute the mean and variance of $z^\text{b}$, then normalize $z^\text{c}$ with these computed values, or \cite{style_aligned} perform AdaIN normalization on $K^\text{b}$ and $K^\text{c}$. Although these direct modifications to $z^\text{c}$ can produce exaggerated image editing effects, they also disrupt semantic details and structural coherence.
    %}

    % Version 2
    %\textcolor{red}{
    In contrast to the aforementioned strategies, we argue that certain attributes crucial to content identity should not be entirely replaced by style features, as discussed in \cite{attributes_and_objects}. For example, the yellow hue of a corgi is an essential part of its identity and should be preserved rather than changed to the global background color tone such as blue or black. Instead, integrating the yellow color from the background style into the corgi would better maintain its content identity as shown in \Figref{fig:attention_strategy}\textcolor{\SubfigColor}{(c)}. To achieve this, we prioritize high-similarity style features that potentially possess content-related attributes such as color, texture, or semantics. Instead of evenly scaling the attention output as in \cite{zstar}, scaling the similarity has a different effect due to the softmax process involved. By scaling the input similarity before applying softmax, high-similarity features are amplified while low-similarity features are diminished in the final attention output. This approach helps minimize content disruption during stylization. Without loss of generality, we place a higher tendency on style reference $z^\text{b}$ by setting $\beta$ to 1.1, and a minor preference on content preservation related to $z^\text{f}$ by setting $\alpha$ to 0.9, our designed \frameworkname{} achieves remarkable painterly harmonization effects without losing content structure and background consistency. The overall algorithm and visualization can be found in Appendix.

    \begin{figure*}[t]
      \centering
      \includegraphics[width=0.9\linewidth]{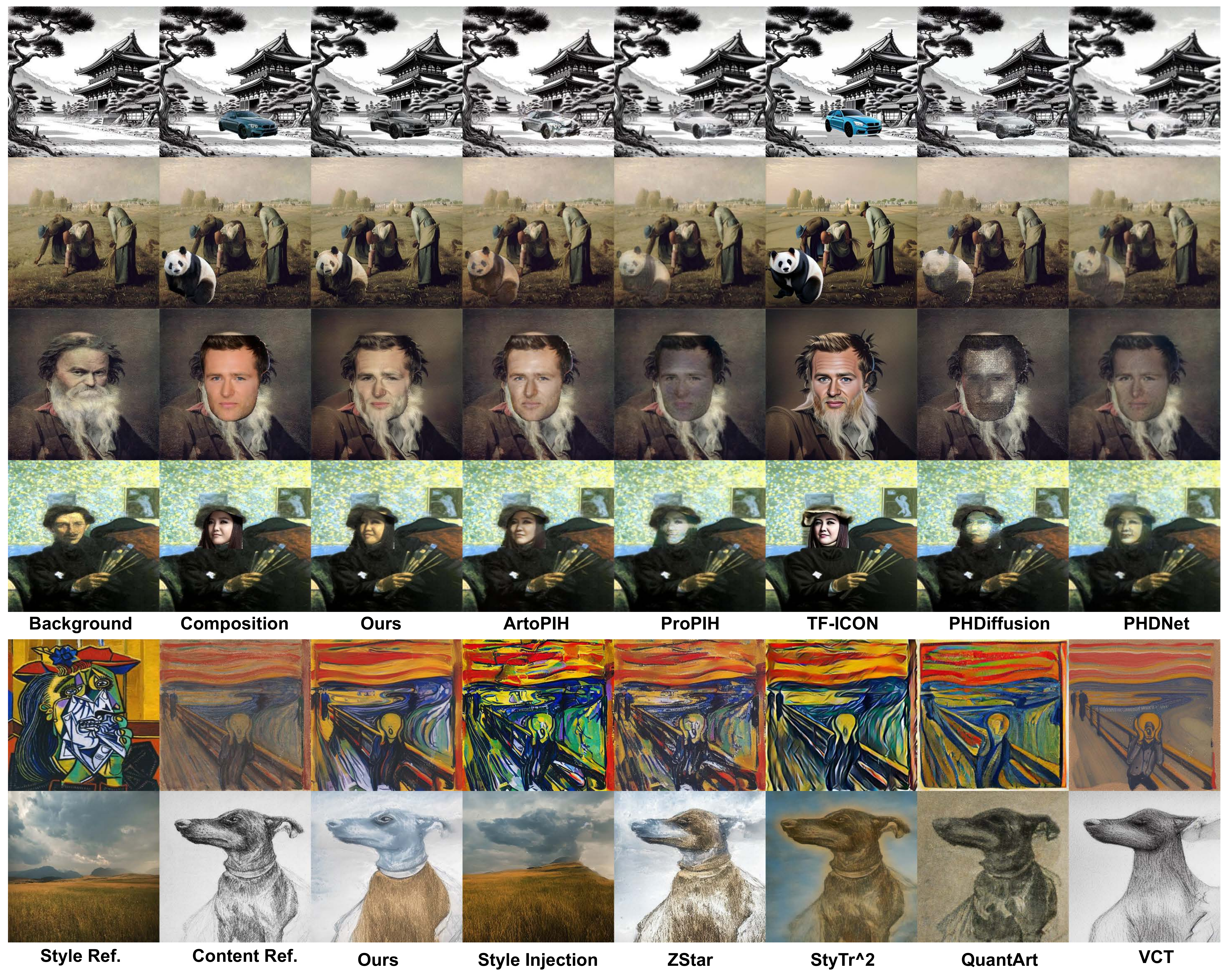}
      \caption{Qualitative result of \textcolor{blue}{object insertion (rows 1 and 2)}, \textcolor{blue}{object swapping (rows 3 and 4)}, and \textcolor{red}{style transfer (rows 5 and 6)}}
     \label{fig:comparison_obj_gph}
     \label{fig:comparison_st_gph}
    \end{figure*}
    %\begin{figure*}[h]
      %\centering
      %\includegraphics[width=0.8\linewidth]{figs/GPH_obj.pdf}
      %\caption{Qualitative result of object insertion (1st and 2nd rows) and object swapping (3rd and 4th rows) on GPH-Benchmark}
     % \label{fig:comparison_obj_gph}
    %\end{figure*}
    
    %\begin{figure*}[h]
    %  \centering
      %\includegraphics[width=0.8\linewidth]{figs/GPH_st.pdf}
      %\caption{Qualitative comparison of style tranfer task on GPH-Benchmark}
      %\label{fig:comparison_st_gph}
    %\end{figure*}
    
    \begin{table*}[t]
      \centering
      \small
      \caption{Quantitative results of GPH-Benchmark (\textsuperscript{\dag} represents the method with inference-time-adjustable hyperparameters. The left side of / represents content emphasis strategy, while the right side of / represents stylized emphasis strategy.)}
      \label{tab:main_gph}
      \begin{tabular}{c|ccccc|cccc} 
        \hline
         & \multicolumn{5}{c|}{Painterly Harmonization (512x512)} & \multicolumn{4}{c}{Style Transfer (512x512)} \\
         \hline 
        & Ours\textsuperscript{\dag} & ArtoPIH & ProPIH\textsuperscript{\dag} & TF-ICON\textsuperscript{\dag} & PHDiff\textsuperscript{\dag} & Ours\textsuperscript{\dag} & StyleID\textsuperscript{\dag} & Z-STAR\textsuperscript{\dag} & StyTr$^2$ \\
        \hline
        $Venue$  & - & AAAI'24 & AAAI'24  & ICCV'23  & MM'23  & - & CVPR'24 & CVPR'24 & CVPR'22 \\
        $LP_\text{bg} \downarrow$  & \textbf{0.11}/0.12 & 0.25 & 0.31/0.31 & 0.20/0.36 & 0.12/0.12 & 0.72/\textbf{0.55} & 0.60/0.56 & 0.70/0.63 & 0.61  \\
        $LP_\text{fg} \downarrow$  & \textbf{0.10}/0.32 & 0.37 & 0.34/0.42 & 0.32/0.36 & 0.10/0.39 & \textbf{0.11}/0.45 & 0.36/0.48 & 0.15/0.37 & 0.40 \\
        $CP_\text{img} \uparrow$ & \textbf{95.42}/78.63 & 84.96 & 87.78/77.24 & 85.35/82.05  & 95.13/73.65 & \textbf{96.43}/69.57 & 83.26/69.80 & 92.31/77.20 & 83.57 \\
        $CP_\text{st} \uparrow$  & 47.50/\textbf{56.37} & 49.67 & 47.87/51.19 & 47.66/47.40 & 47.64/55.96 & 57.97/\textbf{78.60} & 67.70/77.47 & 59.61/69.50 & 63.28\\
        $CP_\text{dir} \uparrow$ & 0.11/11.69 & 5.40 & 2.83/10.09 & 2.96/4.63   & 0.35/\textbf{15.39}  & 3.97/\textbf{51.59} & 26.96/50.24 & 9.76/34.83 & 22.08 \\
        \hline

        \bottomrule 
      \end{tabular}
    \end{table*}

\section{Experiments} \label{sec:exp}
  
    \noindent\textbf{Setup.} We employ the Stable Diffusion model~\cite{LDM} as the pretrained backbone and utilize DPM Solver++ as the scheduler with a total of 25 steps for both inversion and reconstruction. Specifically, we first resize the input images $I^\text{f}$, $I^\text{b}$, and $I^\text{c}$ to 512$\times$512, and encode them into corresponding $z^\text{f}_0$, $z^\text{b}_0$, and $z^\text{c}_0$. Afterward, we take these latents with prompt embedding $\rho_{exceptional}$ as the input during both inversion and reconstruction stage. As for the rest of setting, we refer these hyperparameters ($T_{share}$, $L_{share}$, $\alpha$, $\beta$) as ``inference-time-adjustable hyperparameters'' since they can be flexibly adjusted to modulate the strength of style according to different use cases during the inference process, we leaves remain setting details in the Appendix.
    %Due to the space constraint, please refer to \Appref{app:experiment_hyper} for the setting of the hyperparameters and sensitivity test.
    %For the style transfer task, we set $\alpha$ to 0.9, $\beta$ to 1.1, $T_{share}$ to 25 (we activate the share-attention layer when t<$T_{share}$), and $L_{share}$ to 14 (out of 16 total layers in the diffusion U-Net). As for the object swapping purpose, we change $T_{share}$ to 20, and for the object insertion usage, we change $T_{share}$ to 15. 
    %\textbf{\frameworkname setting.}
    %\noident\textbf{baselines.}
    
   \noindent\textbf{Datasets.} We generalize the computational metrics and benchmarks from various image editing methods including ``Painterly image harmonization'', ``Prompt-based Image Composition'', and ``Style Transfer''. Additionally, we examined different approaches on our proposed ``General Painterly Harmonization'' achieved by the General Painterly Harmonization Benchmark (GPH Benchmark). This benchmark generalizes real-world usage scenarios of the aforementioned methods including ``Object Insertion'', ``Object Swapping'' and ``Style Transfer'', providing a more practical benchmark and aims to mitigate the shortcomings of existing benchmarks such as WikiArt combined with COCO \cite{wikiart, COCO} and the TF-ICON Benchmark \cite{tficon}. Details and experiment of these datasets can be found in the Appendix.
  
    \noindent\textbf{LPIPS and CLIP regarding computation metrics.} In our evaluation, we use LPIPS \cite{LPIPS} and CLIP \cite{CLIP} metrics, abbreviated as $LP$ and $CP$ respectively. LPIPS is sensitive to low-level visual features, while CLIP excels in capturing high-level semantic features. In TF-ICON benchmark, these two metrics are leveraged to assess content preservation and stylized performance, where $LP_\text{fg}$ and $CP_\text{img}$ are calculated to measure the content consistency and image semantic similarity, respectively. Moreover, $LP_\text{bg}$ is also used to measure background consistency before and after harmonization. And $CP_\text{dir}$ \cite{stylegan-nada} to calculate the alignment level between the feature shift direction of pasted object and the background. Finally, we adopt $CP_\text{st}$ \cite{VCT} to measure the feature similarity of harmonized images and style references. 
    
    \noindent\textbf{Range-based evaluation.} While metrics such as LPIPS and CLIP are useful for assessing content fidelity and stylization in image harmonization, they can sometimes \textbf{emphasize either too much content preservation or excessive stylization}, resulting inharmonious image. Therefore, we argue that an effective pipeline should offer users the flexibility of balancing between stylization intensity and content integrity by adjusting hyperparameters. To measure this capability, we suggest defining upper and lower bounds for content preservation and stylization, which can serve as indicators of a method’s adaptability across different harmonization scenarios. The corresponding upper and lower settings for the baselines are provided in the Appendix.

    \noindent\textbf{Baselines.} We compared our proposed \frameworkname{} with different state-of-the-art methods on various tasks for the comprehensive assessment. For ``Painterly Image Harmonization'', we incorporate ArtoPIH \cite{artopih}, ProPIH \cite{propih}, PHDiffusion \cite{PHDiff}, and PHDNet \cite{PHDNet}. For ``Prompt-Based Image Composition'', we use TF-ICON \cite{tficon}. In the ``Style Transfer'' category, we evaluate Style Injection (StyleID) \cite{style_injection}, ZSTAR \cite{zstar}, StyTr2 \cite{StyTr2}, QuantArt \cite{QuantArt}, and VCT \cite{VCT}. For models designed for 256x256 resolutions, e.g. ProPIH, we resized the output to 512x512 for high-resolution evaluation. Comparisons of 256x256 resolution are in the Appendix.

  \subsection{Qualitative Comparison}
  For qualitative comparison, \frameworkname{} showcases remarkable capabilities in our proposed GPH Benchmark as depicted in \Figref{fig:comparison_obj_gph}, ranging from low-level texture harmonization such as transitioning to singular colors and color matching (rows 1 and 2) to high-level semantic harmonization such as extending the skin color of the replaced man onto the pasted face or redrawing the covered beard along with the chin line of the pasted face (rows 3 and 4). Although ArtoPIH and ProPIH are able to achieve low-level texture harmonization, they struggle with high-level semantic blending, such as face recovery in the row 3 of \Figref{fig:comparison_obj_gph}, due to training data limitation. This highlights how our similarity reweighting technique effectively leverages the characteristics of the diffusion model to achieve both texture and semantic harmonization with image-wise attention.
  
  Moreover, our proposed \frameworkname{} demonstrates exceptional performance in style transfer (row 5,6 in \Figref{fig:comparison_st_gph}). Our method outperforms others in stylizing original content while maintaining high image quality, effectively mitigating the common issue of content disruption seen in other attention-based methods. For instance, our approach preserves content coherence more accurately than StyleID and ZSTAR as shown in row 5. Furthermore, our model excels in blending photographic features into sketches, where other methods fail, as depicted in the row 6. This illustrates that our similarity disentangle mask not only preserves content information effectively but also extracts style features robustly, even in scenarios like photography.% Additional qualitative results are available in \Appref{app:qualitative}.

  \subsection{Quantitative Results}
  \Tabref{tab:main_gph} presents the quantitative results of the GPH Benchmark. The performance of \frameworkname{} consistently surpasses that of existing evaluation criteria on different benchmarks. Our similarity disentangle mask significantly improves reference preservation compared to prompt-based editing methods such as TF-ICON, as well as traditional harmonization methods like ArtoPIH and ProPIH, achieving the lowest $LP_\text{bg}$ and $LP_\text{fg}$ values while also demonstrating superior stylization with the highest $CP_\text{st}$.
  
  Moreover, \frameworkname{} employs a novel similarity-based editing technique that consistently outperforms existing attention-based methods, such as StyleID, in both content preservation metrics ($LP_\text{fg}$, $CP_\text{img}$) and stylization metrics ($CP_\text{st}$, $CP_\text{dir}$). Additionally, the wide content preservation and stylization range of \frameworkname{} confirm the potential of our inference-time-adjustable hyperparameters, which can accommodate various preferences.
  
\begin{figure}[t]
      \centering
      \includegraphics[width=0.9\linewidth]{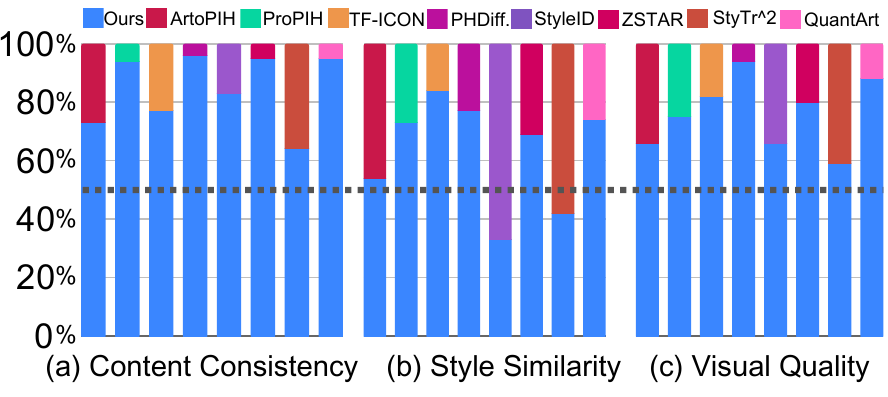}
      \caption{User preferencestudy result.}
      \label{fig:user_hist}
    \end{figure}

  We also conduct user preference studies, which are viewed more reliable \cite{sdxl}. The study encompasses two tasks: Style Transfer and Painterly Harmonization, which includes object insertion and swapping. For each task, we recruited 20 participants, each asked with responding to 20 image pairs. Participants were instructed to compared the generated images based on three criteria: (1) Content Consistency, (2) Style Similarity, and (3) Visual Quality. We provide the results in \Figref{fig:user_hist}, where \frameworkname{} achieving the highest preference in overall quality and content consistency, along with competitive style similarity. These results validate our hypothesis that visual quality transcends mere content preservation or style strength.% More details in \Appref{app:user_study}.

    % \begin{figure}
    %   \centering
    %   \includegraphics[width=\linewidth]{figs/comparison_tf_benchmark.png}
    %   \caption{Results on TF-Benchmark. Our results (second row) show better content preservation than TF-ICON (first row).}
    %   \label{fig:comparison_TF}
    % \end{figure}

    %\noindent\textbf{\frameworkname{} component.} 
   \subsection{Ablation Study}
   \Tabref{tab:ablation} reveals the impact of components within \frameworkname{}. Simply applying reconstruction to the composite image is ineffective at harmonizing pasted object into background. By contrast, when we incorporating similarity disentangle mask, we perfectly disentangle the attention of $z^\text{c}$ to the two other image latents $z^\text{f}$, $z^\text{b}$ and reach \textbf{nearly no reconstruction loss}. Furthermore, the integration of the similarity reweighting strategy significantly improves the stylization indices $CP_\text{st}$ and $CP_\text{dir}$ across both tasks, demonstrating its effectiveness in encoding cross-image information into the composite image.
   
   %\noindent\textbf{Inference-time-adjustable Hyperparameters.} Here, we validate the critical role of flexibility in image editing applications. In \frameworkname{}, the hyperparameters $\alpha$ and $\beta$ directly influence the stylization strength, while $T_{share}$ and $L_{share}$ are pivotal in the harmonization process. Specifically, higher values of $\alpha$ and $\beta$, and lower values of $T_{share}$ and $L_{share}$, help to align the output image more closely with the style reference. These adjustments afford \frameworkname{} enhanced adaptability across a diverse array of usage scenarios. For a more detailed discussion, please refer to \Appref{app:experiment}.
   
   %\noindent\textbf{Extension of Share-Attention Layer} As the share-attention layer is designed for more general purposes of image editing methods, we explore the potential of share-attention layer as a hot-plug-in component of other image editing utilities such as inpainting and image synthesis in \Appref{app:more_applications}.

    \begin{table}
      \centering
      \caption{Ablation study on \frameworkname{}'s components in painterly harmonization (upper) and style transfer (bottom) on GPH Benchmark. (We abbreviate ``Similarity Disentangle Mask'' and ``Similarity Reweighting'' as ``SDM'' and ``SR'' ). Because the +SDM+SR is the stylization emphasis strategy of +SDM, we put its stylization upper bound here.}
      \label{tab:ablation}
      \begin{tabular}{l|cccc} 
        \toprule 
        Metrics & $LP_\text{bg} \downarrow$  & $LP_\text{fg} \downarrow$ &  $CP_\text{st} \uparrow$  & $CP_\text{dir} \uparrow$ \\
        \midrule
        Reconstruction & \textbf{0.11}  & \textbf{0.09}  &  47.50  & 0.11 \\
        +$SDM$ & 0.11  & 0.10  &  47.50  & 0.18 \\
        +$SDM$+$SR$ & /0.12  & /0.32  & /\textbf{56.37}  & /\textbf{11.69} \\
        \midrule
        \midrule
        Reconstruction & 0.72  & \textbf{0.11}  &  57.97  & 3.97 \\
        +$SDM$ & 0.69 & 0.12  &  59.05  & 5.12 \\
        +$SDM$+$SR$ & /\textbf{0.56}  & /0.45  & /\textbf{78.60}  & /\textbf{51.59} \\
        \bottomrule 
      \end{tabular}
    \end{table}
 
    %\begin{table}
    %  \centering
    %  \caption{Ablation study on \frameworkname components in Style Transfer}
    %  \label{tab:ablation_style}
    %  \begin{tabular}{c|ccccc|} 
    %    \toprule 
    %    Metrics & $LP_\text{bg} \downarrow$  & $LP_\text{fg} \downarrow$ & $CP_\text{img} \uparrow$ &  $CP_\text{style} \uparrow$  & $CP_\text{dir} \uparrow$ \\

    %    \midrule
    %    Reconstruct & 0.72  & 0.11 & 96.43 &  57.97  & 3.97 \\
    %    $+SA$ & 0.70 & 0.11 & 96.15 &  58.56  & 4.84 \\
    %    +$Reweighting$ & 0.63  & 0.33 & 83.84 & 66.68  & 25.39 \\
    %    \bottomrule 
    %  \end{tabular}
    %\end{table}

  \section{Conclusion}
  In this work, we introduce a novel \textbf{similarity disentangle mask}, faciliatating the utilization of attention from different images. Furthermore, we devised the \textbf{similarity reweighting} technique capable of controlling the attention strength of reference images without the need for fine-tuning or prompt. Based on them, we propose \textbf{\frameworkname{}} to perform a more general form of painterly harmonization. Also, we construct the \textbf{GPH Benchmark} with \textbf{range-based evaluation} aiming to mitigate the current shortage of evaluations for image editing. Both human and quantitative evaluations show that \frameworkname{} produces more harmonious results, which should benefit future research in image editing.

\section*{Acknowledgments}
  This work is partially supported by the National Science and Technology Council, Taiwan under Grants NSTC-112-2221-E-A49-059-MY3 and NSTC-112-2221-E-A49-094-MY3.

\bibliography{aaai25}

\clearpage
\appendix
\section{Appendix: Algorithm}\label{app:algorithm}
    The \frameworkname{} framework is based on stable-diffusion (SD)~\cite{LDM}, combined with DPM-solver~\cite{dpm_solver}, which not only reduces the timestep requirements but also supports the functionality of the inversion process. And this inversion process can be further stablized with a fixed prompt embedding $\rho_{\text{except}}$.  We assume that the inversion process for the \frameworkname{} input has been completed. In the subsequent reconstruction stage, cross-image information is incorporated into the output image via our share-attention module. Our proposed share-attention module is a plug-and-play component, designed to replace the attention layer in the original SD framework. Furthermore, Our focus lies in the share-attention layer's forward function (share-FORWARD) with additional similarity disentangle mask $\hat{M}$, where we only substitute the forward mechanism of the original attention layer while retaining trained parameters (Q-K-V Projection layer, normalization layer, etc.). The detailed algorithm for \frameworkname{} is outlined in \Algref{alg:overall}.
    \begin{algorithm}[]
    \SetAlgoLined
    \SetKwComment{Comment}{$\triangleright \ $}{ }
    \KwData{initial noise $Z_T=cancat\{z_T^{\text{f}}, z_T^{\text{b}}, z_T^{\text{c}}\}$, step $T_{\text{share}}$ and layer depth $L_{\text{share}}$ to start using share attention module, the similarity weight $\hat{\alpha}$ and $\hat{\beta}$}
    %exceptional prompt embedding $\rho_{\text{except}}$, null prompt embedding $\varnothing$ , CFG scale $gs$

    \KwResult{Harmonized $Z_0$}
    %\For{$t=T, T-1,...,1$}{
    %    $N_{uncond}\leftarrow$ FORWARD($Z_t$,$\varnothing$,$t$)\;
    %    $N_{cond} \leftarrow$ FORWARD($Z_t$,$\rho_{\text{except}}$,$t$)\;
    %    $N_{pred} \leftarrow$ $N_{uncond}+gs\cdot(N_{cond}-N_{uncond})$\;
    %    $Z_{t-1} \leftarrow DPMSolver(Z_{t},N_{pred},t)$
    %}
    \Comment{\textcolor{red}{We use default stable diffusion model with exceptional prompt embedding $\rho_{\text{exceptional}}$ as input, while rewriting the FORWARD of attention layer to Share-FORWARD}}
    \Comment{\textcolor{red}{We omit the linear transform and layer normalization, for brevity}}

    \textbf{Share-FORWARD}($Z_t$,$C$,$\hat{\alpha}$,$\hat{\beta}$,$t$):\\
        \Indp$O_0 \leftarrow Z_t$\;
        \For{$l=0,1...L$}{
            $\hat{Q},\hat{K},\hat{V} \leftarrow$ $Proj_l$($O_l$)\;
            %$\hat{Q},\hat{K},\hat{V} \leftarrow$ reshape($Q,K,V$)\;
            \eIf{$t<T_{share}$ and $l>L_{share}$}
            {
                \Comment{\textcolor{comment_green}{Start image-wise attention with reweighting}}
                set ($\alpha$,$\beta$, $\gamma$) in $\hat{M}$ to ($\hat{\alpha}$, $\hat{\beta}$, $-\infty$)}
            {
                \Comment{\textcolor{comment_green}{Equivalent to normal diffusion process but in different shape}}
                set ($\alpha$,$\beta$, $\gamma$) in $\hat{M}$ to ($-\infty$, $-\infty$, 1)
            }
            $\hat{A}=\text{Softmax}\left(\frac{\hat{M}\odot (\hat{Q}\hat{K}^T)}{\sqrt{d}}\right)\hat{V}$\;
            
            $O_{l} \leftarrow $ $O_l$+$\hat{A}$\;
            \Comment{\textcolor{comment_green}{$CA_l$ is the cross-attention layer at layer l, $C$ is the corresponding text embedding (fixed to $\rho_\text{exceptional}$)}}
            $O_{l+1} \leftarrow $ $O_l$+$CA_l$($O_l$,C) \;
        }
        \textbf{return} $O_L$ \;
    \caption{Training-and-prompt free General Painterly Harmonization}
    \label{alg:overall}
    \end{algorithm}

\section{Appendix: Visualization}
We visualize how the similarity reweighting technique change the attention of $z^\text{c}$ toward two different sources $z^\text{f}$ and $z^\text{b}$. We perform the visualization of attention on layer 14 of UNet in Fig.\textcolor{red}{\ref{fig:attn_vis}} below. With the similarity disentangle mask only (columns 1, 2), the information from foreground dominate the attention of inserted object, when the similarity reweighting is included (columns 3, 4), the background information significantly influences the inserted object from the early denoising step t=15 till the last step t=0.

\begin{figure}[h]
  \centering
    \includegraphics[width=\linewidth]{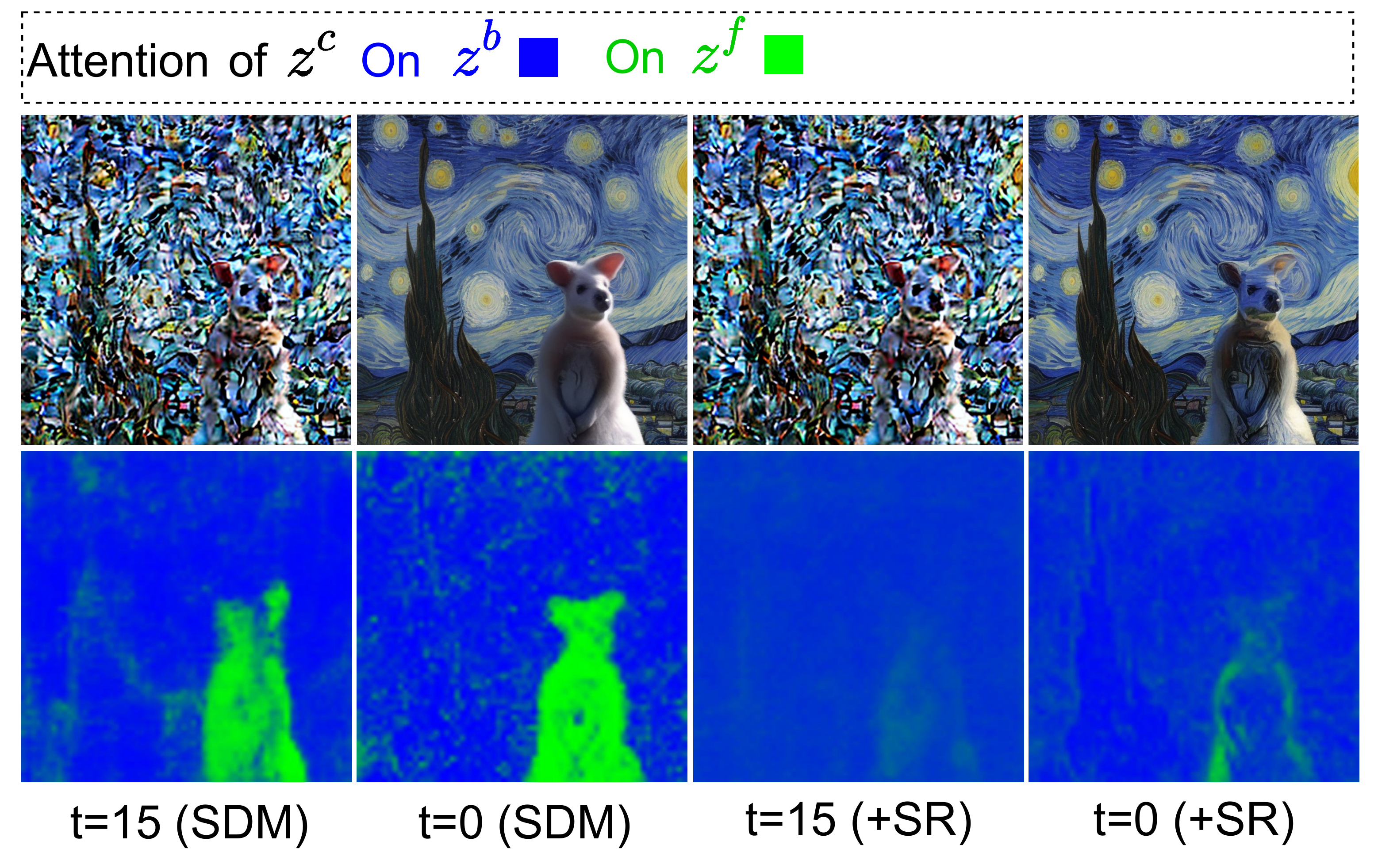}
  % \caption{The visualization of shared-attention layer. The \textcolor{blue}{blue} part denote the attention on background image, while the \textcolor{green}{green} part denote the attention on foreground image.}
  \caption{The visualization of how simialrty reweighting change the attention.}
  \label{fig:attn_vis}
\end{figure}
\begin{figure}[h]
  \centering
    \includegraphics[width=\linewidth]{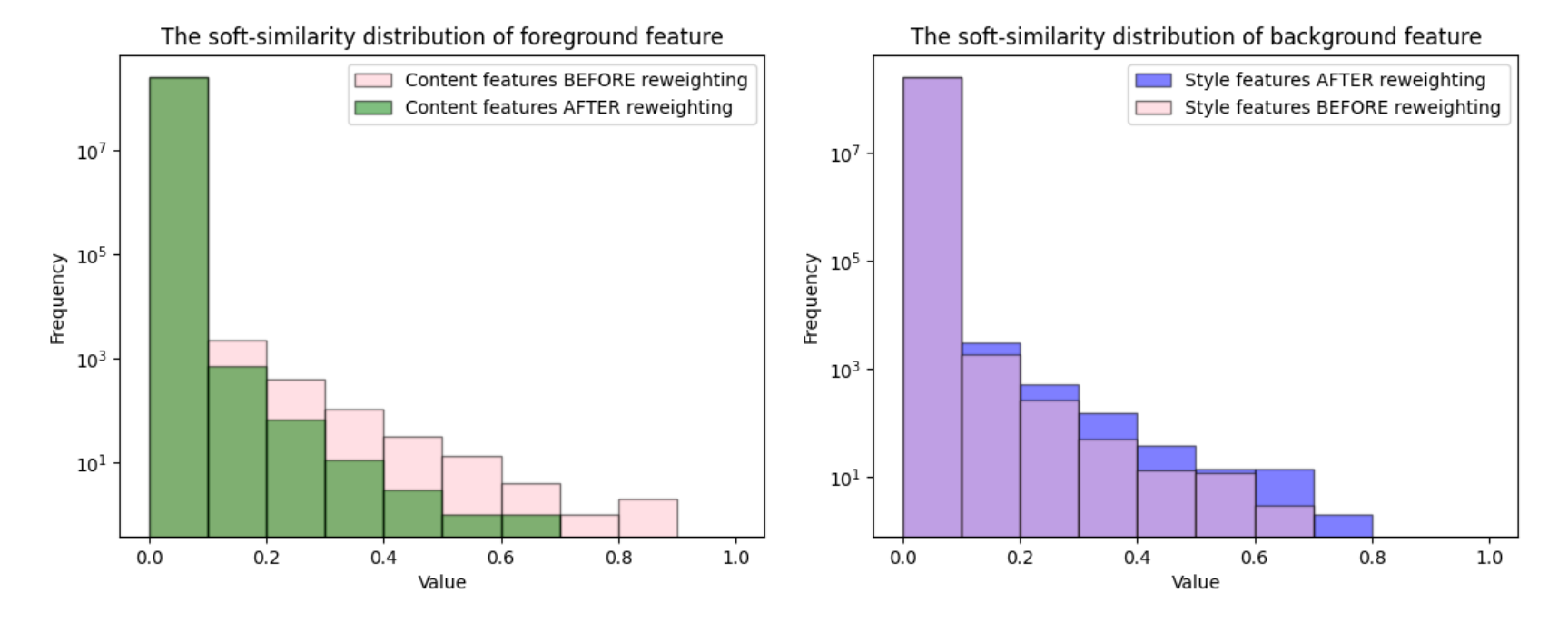}
  % \caption{The visualization of shared-attention layer. The \textcolor{blue}{blue} part denote the attention on background image, while the \textcolor{green}{green} part denote the attention on foreground image.}
  \caption{The visualization of how simialrty reweighting change the attention distribution.}
  \label{fig:dist_vis}
\end{figure}

We also visualize the impact of our similarity reweighting technique on the feature similarity distribution between the composite image and the content and style references, as illustrated in \Figref{fig:dist_vis}. Our first observation is that similarity reweighting effectively reduces the influence of the content reference, thereby creating more space for stylization. Furthermore, we validate our claim that reweighting the similarity before the softmax operation significantly enhances the influence of high-similarity features while diminishing that of low-similarity features. As shown in the right part of \Figref{fig:dist_vis}, although style features predominantly exhibit low similarity values (less than 0.1), the increase in similarity is relatively greater for high-similarity features (values above 0.2).

\section{Appendix: Datasets}\label{app:datasets}
  %We include the latter because our proposed methods can achieve similar effects to prompt-based image composition even without prompt assistance. Additionally, we will consider general style transfer methods \cite{QuantArt, VCT, StyTr2}. To simulate the painterly image harmornization for style transfer approach, we first stylize the composite image and then copy and paste the foreground object onto the background image. %(The primary drawback of this replacement is the neglect of edge interaction between the foreground and background)

    \subsection{WikiArt~\cite{wikiart} with COCO~\cite{COCO}.}
    This dataset has been widely adopted by various general stylization methods \cite{PHDiff, PHDNet, QuantArt, StyTr2} due to its high flexibility and feasibility. Following common evaluation practices, we utilize the WikiArt dataset for background images and the COCO dataset as the source for foreground objects. Specifically, we randomly sampled 1000 images from the WikiArt validation dataset and 1000 segmented objects across 80 different classes from the COCO validation dataset (with each object's class equally distributed). These segmented objects are then composited onto the background images to generate our final composite images. (The evaluation result can be found in \Figref{fig:comparison_wikiart}, \Tabref{tab:main_coco512} and \Tabref{tab:main_coco256}).However, a limitation of this dataset lies in its diversity, which is constrained to combinations of real-world objects (from COCO) against paintings (typically European style). As a result, the data distribution from WikiArt combined with COCO may not fully represent real-world applications of image harmonization tasks, which often involve combining fictional objects with various forms of graphic art. %However, the flexibility of this approach, while beneficial, also makes it difficult to regulate, due to the over-whelming possible combinations. %Therefore, previous research has not yet reached a consensus on the evaluation metrics nor the restrict rules for this dataset. Therefore, we will borrow the computation metrics from our proposed GPH Benchmark, which will be discussed in the latter section.
    
    \subsection{TF-ICON Benchmark~\cite{tficon}.}
    This dataset was originally designed for prompt-based image composition, with each data entry containing four components: a text-generated background image, a reference image of a real-world object, a composite image of the real-world object with the background, and a prompt describing the composite image. The background images encompass four visual domains: cartoon, photorealism, pencil sketching, and oil painting. For evaluate the prompt-free ability of our proposed \frameworkname{} method, \textbf{we omit the given text descriptions and only use the images as input}. The evaluation result can be found in \Figref{fig:comparison_TF} and \Tabref{tab:tf_benchmark}. While this benchmark provides an additional prompt for more flexible evaluation, it lacks diversity as the backgrounds are all images produced by a generative model for the purpose of aligning prompt description to background style, lacking data of real-world paintings such as famous paintings "Starry Night," which are widely adopted in real-world stylized applications.
    
    \subsection{General Painterly Harmonization Benchmark.}
    We have observed the drawbacks of the aforementioned dataset--lacking strong correlation toward the usage of image composition related tasks in real-world applications. Hence, we propose the General Painterly Harmonization Benchmark (GPH-Benchmark), aiming to solve not only the generalizability issues of existing datasets but also the shortcomings of current evaluation metrics by computing the content/stylized range as an approximation of harmonization ability. Beginning with the construction of the dataset, our objective is to generalize three main applications commonly used by human users in real-world scenarios: \textbf{object swapping}, \textbf{object insertion}, and \textbf{style transfer}. Our dataset comprises source data from real-world objects, generated objects, famous painterly backgrounds, and generated backgrounds with unique styles, resulting in a total of 635 test cases that cover various examples of general painterly harmonization created by human labor. We partition the conventional painterly image harmonization task into two distinct subcategories: object swapping and object insertion. The primary distinction lies in the objectives pursued by each subcategory. Object swapping aims for high-level semantic harmonization, emphasizing strong semantic connections between the swapped object and the background image. For instance, in the case of swapping faces, the goal is to ensure natural integration with surrounding features like hair and skin color. On the other hand, object insertion prioritizes low-level visual naturalness, focusing on harmonizing edges and textures to achieve visual coherence.

\section{Appendix: Baselines }\label{app:baselines}

  \textbf{ProPIH \cite{propih}}:
  ProPIH is a novel painterly harmonization pipeline, different from previous autuencoder-based methods, they design a multi-stage harmonization network, which harmonize the composition foreground from low-level style to high-level style. We directly choose the first stage output as the stylization lower bound, while the last stage output as stylization output.

  \noindent\textbf{PHDiffusion \cite{PHDiff}}:
  PHDiffusion (abbreviated as PHDiff in subsequent sections) is a framework for painterly harmonization. They propose incorporating an additional adaptive encoder combined with a fusion module into the existing stable diffusion pipeline and fine-tuning this combined pipeline on the WikiArt with COCO dataset. It is noteworthy that within this framework, they introduce an additional "Strength" hyperparameter to control the scale of the fusion module, which also represents the influence of the background style on the pasted object. Consequently, to evaluate the performance of the content emphasis strategy of PHDiff, we set the 'Strength' parameter to 0, while for the stylization emphasis strategy, we set it to 1. (Their default setting for 'Strength' is fixed at 0.7).

  \noindent\textbf{TF-ICON \cite{tficon}}:
  The TF-ICON approach primarily leverages two hyperparameters to govern the prompt-based image composition process. Firstly, $\tau_\alpha$ indicates the onset of attention injection (0 for beginning, 1 for the end), and secondly, $\tau_\beta$ denotes the onset of the rectification process. Unfortunately, neither $\tau_\alpha$ nor $\tau_\beta$ directly modulates stylization intensity. However, reducing $\tau_\alpha$ generally yields more stylized outputs, while reducing $\tau_\beta$ produces images closer to the composite. Thus, to assess TF-ICON's content emphasis strategy, we follow the setting suggest by TF-ICON, we set $\tau_\alpha=0.4$ and $\tau_\beta=0$ as proposed in their paper for photography composition, demanding heightened content preservation. Conversely, for the stylization emphasis strategy, we adopt $\tau_\alpha=0.4$ and $\tau_\beta=0.8$ as recommended for cross-domain composition, necessitating higher stylization strength.

  \noindent\textbf{ZSTAR \cite{zstar}}:
  They reveal that the cross-attention mechanism in latent diffusion models tends to blend the content and style images, resulting in stylized outputs that deviate from the original content image. To overcome this issue, they introduce a cross-attention rearrangement strategy, for stlization lowe bound we restrict this rearrangement only to the middle 16th attention layer, as for stylization upper bound, we allow these arrangement in all the 1st to 32th attention layer.
  
  \noindent\textbf{StyleID \cite{PHDiff}}: Furthermore they introduce query preservation and attention temperature scaling to mitigate the issue of disruption of original content and initial latent Adaptive Instance Normalization (AdaIN) to deal with the disharmonious color, they already provide the default settings, for stylization lower bound they recommend setting gamma to 0.75. As for stylization upper bound, they suggest the setting of gamma to 0.3.

\section{Appendix: More Qualitative Result }\label{app:qualitative}
  \subsection{Qualitative of WikiArt w/ COCO}\label{app:qualitative_wikiart}
We compare our proposed method with common WikiArt w/ COCO baselines in \Figref{fig:comparison_wikiart}. As shown, our method produces convincing outputs across a wide range of styles, including combinations of objects such as humans, food, and animals merged with styles like Impressionism, Modern sketch, and Abstract painting. Unlike other baselines that primarily focus on matching the color tones of objects with the background, our method better utilizes the existing elements of the background reference, such as brushstrokes, abstract edges, and inherent colors. This results in more coherent and harmonious outputs, outperforming existing models.

  \subsection{Qualitative of TF-Benchmark}  \label{app:qualitative_tficon}
\begin{figure}[h]
\centering
\includegraphics[width=\linewidth]{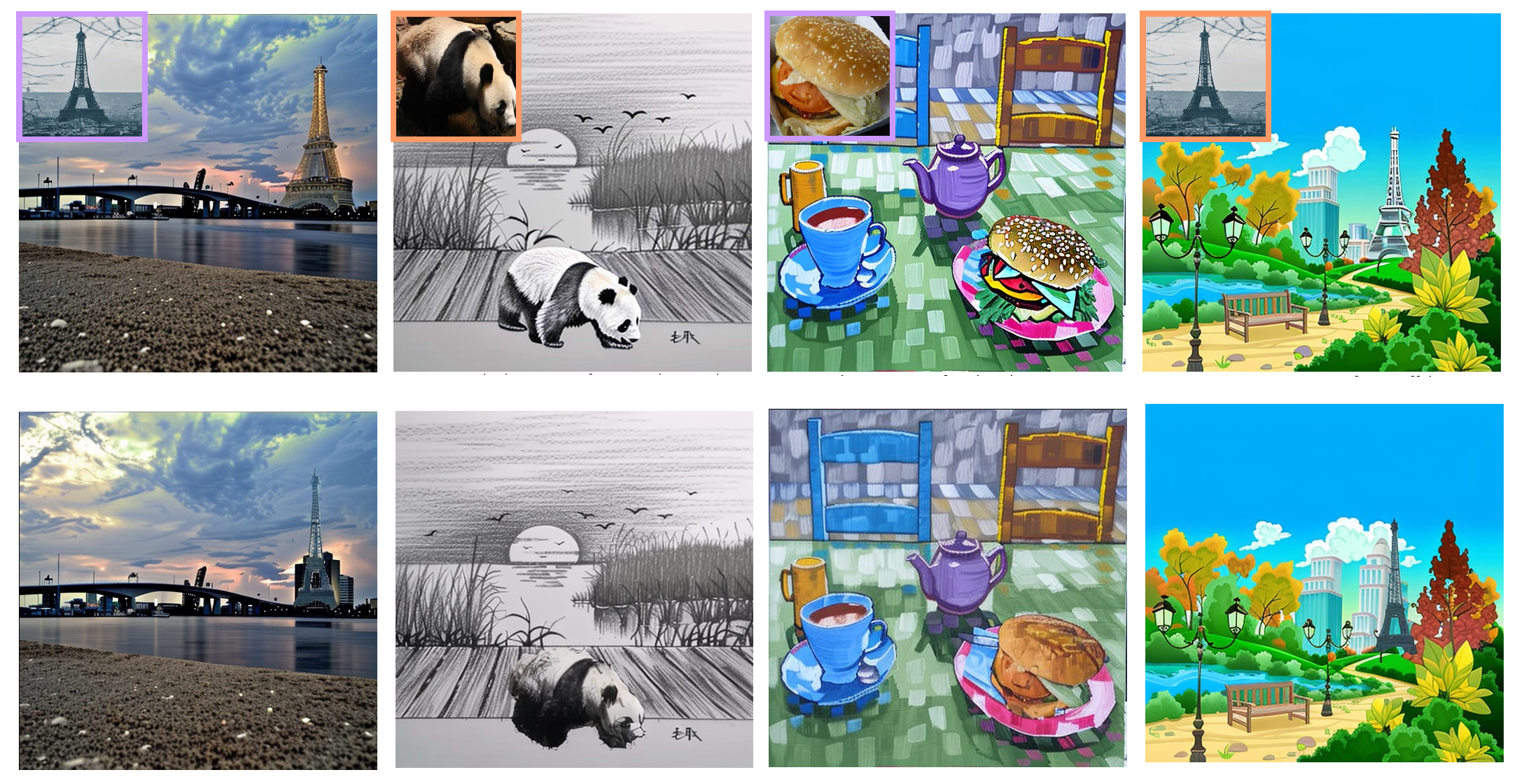}
\caption{Results on TF-Benchmark. Our results (second row) show better content preservation than TF-ICON (first row).}
\label{fig:comparison_TF}
\end{figure}

  We provide the comparison of our proposed method toward other baselines in \Figref{fig:comparison_TF}. Although TF-ICON produces harmonious outputs, the content often becomes distorted. For instance, the panda’s pose is changed, the hamburger’s content is altered, and the tower’s shape is modified. In contrast, our method not only better preserves the identity of the objects but also seamlessly blends them into the reference background. Showing the advantage of TF-GPH in the aspect of content identity preservation.

  \subsection{More Qualitative result of GPH Benchmark}
  We provide more result of \frameworkname{} on GPH Benchmark in \Figref{fig:more_gph_mix} (Insertion and swapping) and \Figref{fig:more_gph_style} (Style Transfer). These example contains novel objects which are not included in the common COCO dataset, such as pyramid, cartton character, and seal. We also include the sumi-e, cartoon and menga background reference serving as content or style references as shown in the last 3 row in \Figref{fig:more_gph_style} (Style Transfer). These examples validate the efficacy of TF-GPH methods upon uncommon input, which are often out of the training data of common painterly harmonization dataset.

\section{Appendix: More Quantitative Result }\label{app:quantitative}

  \subsection{Quantitaive of GPH Benchmark in 256x256}\label{app:quantitative_wikiart}
  Our proposed TF-GPH is a novel pipeline designed for generating images at resolutions of 512x512 and higher. In contrast, existing methods like ArtoPIH and ProPIH are limited to generating images at 256x256 resolution. To thoroughly evaluate the performance of these models in low-resolution scenarios, we resize the output of our methods to facilitate painterly harmonization at lower resolutions \Tabref{tab:main_gph}. As shown in the table, our TF-GPH method still produce competative result especially in the stylization related metrics.
  
  \subsection{Quantitaive of WikiArt w/ COCO}\label{app:quantitative_wikiart}
  We provide the comparison of range-based evaluation on WikiArt combined with COCO in \Tabref{tab:main_coco512} (512x512) and \Tabref{tab:main_coco256} (256x256) . 

\subsection{Quantitative Evaluation on TF-ICON Benchmark} We also evaluate the performance of our proposed TF-GPH on the existing prompt-guided image composition benchmark, TF-ICON. As shown in \Tabref{tab:tf_benchmark}
, our method achieves state-of-the-art performance on this benchmark, even without the support of prompts.

\begin{table}
  \centering
  \caption{Quantitative result of TF-ICON Benchmark}
  \label{tab:tf_benchmark}
  \begin{tabular}{l|cccc} 
    \toprule 
     Method & $LP_\text{bg} \downarrow$ & $LP_\text{fg} \downarrow$ & $CP_\text{img} \uparrow$ & $CP_{text} \uparrow$ \\
    \midrule
     SDEdit (0.4) & 0.35 & 0.62 & 80.56 & 27.73\\
     % SDEdit (0.6) & 0.42 & 0.66 & 77.68 & 27.98\\
     Blended & 0.11 & 0.77 & 73.25 & 25.19 \\
     Paint & 0.13 & 0.73 & 80.26 & 25.92 \\
     DIB & 0.11 & 0.63 & 77.57 & 26.84 \\
     TF-ICON & 0.10 & 0.60 & 82.86 & 28.11 \\
     Ours & \textbf{0.05} & \textbf{0.48} & \textbf{83.34} & \textbf{30.33} \\
    \bottomrule 
  \end{tabular}
\end{table}

\section{Appendix: User Study}\label{app:user_study}
\begin{figure}[h]
\centering
\includegraphics[width=\linewidth]{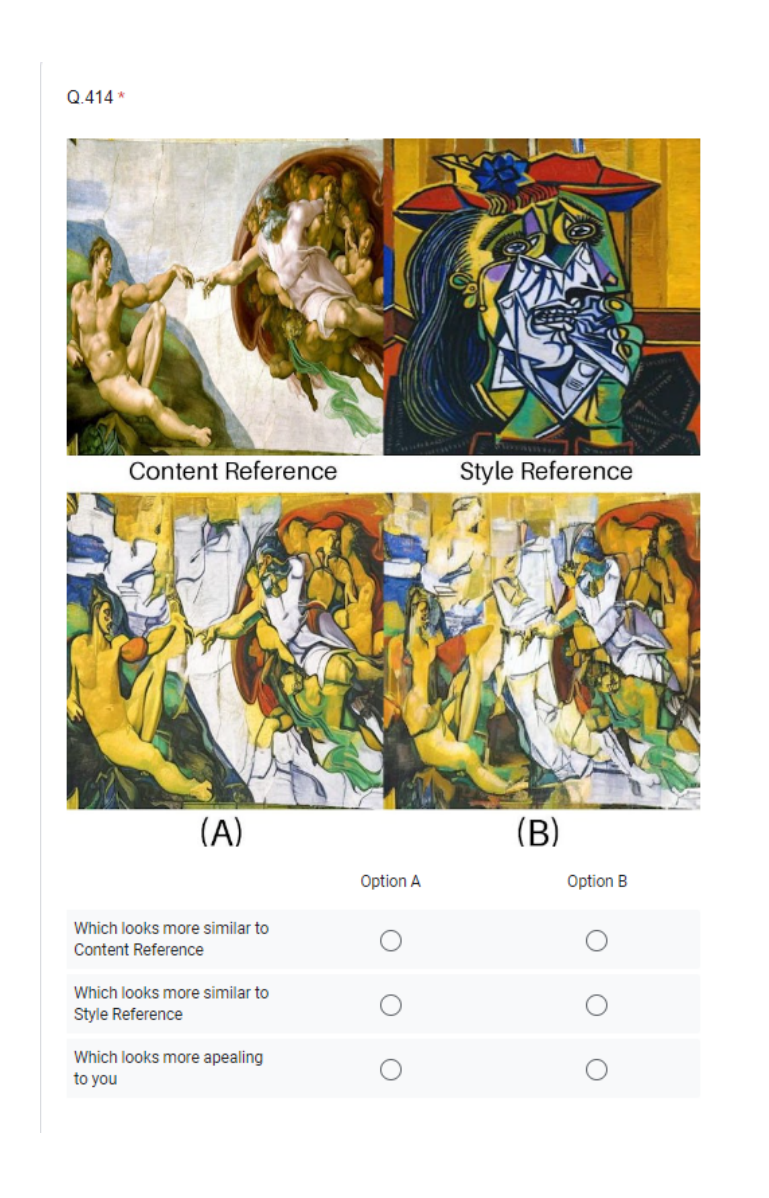}
\caption{A painterly harmonization question participant might need to answer.}
\label{fig:question_demo}
\end{figure}

\textbf{Survey flow.}
At the outset, each participant will receive instructions and a demonstration question aimed at familiarizing them with the answer flow and evaluation criteria ("Content Consistency," "Style Similarity," "Visual Quality"). Subsequently, they will be required to answer 20 randomly selected questions from a pool of 60 questions, along with an attention-check question designed to assess the validity of their responses (details provided in the following section). Furthermore, the options for each question will be shuffled for enhanced reliability. The participant's view of the question is illustrated in \Figref{fig:question_demo}.

\section{Appendix:  Inference-time-adjustable Hyperparameters}\label{app:experiment}

    \subsection{Experement Seeting}\label{app:experiment_hyper}
     For the style transfer task, we set $\alpha$ to 0.9, $\beta$ to 1.1, $T_{share}$ to 25 (we activate the share-attention layer when t<$T_{share}$), and $L_{share}$ to 14 (out of 16 total layers in the diffusion U-Net). As for the object swapping purpose, we change $T_{share}$ to 20, and for the object insertion usage, we change $T_{share}$ to 15. We refer to these hyperparameters ($T_{share}$, $L_{share}$, $\alpha$, $\beta$) as ``inference-time-adjustable hyperparameters'' since they can be flexibly adjusted to modulate the strength of style according to different use cases during the inference process. As we have emphasized, there is no universally optimal harmonization sweet point for all aesthetic preferences; it depends entirely on the specific use case. The only quantitative metrics we can establish are the lower and upper bounds of stylization. Therefore, instead of exhaustively evaluating every possible combination of hyperparameters, we selected settings that produce relatively harmonious outputs for our preference.
    
    \subsection{Sensitive Test}\label{app:experiment_hyper}
    The \frameworkname{} incorporates four adjustable hyperparameters during inference: $T_\text{share}$, $T_\text{L}$, $\alpha$, and $\beta$. The initial two parameters, $T_\text{share}$ and $T_\text{L}$, govern the commencement timing of the share-attention layer; an earlier start of the share-attention layer leads to increased blending of objects into the background. Meanwhile, the latter two parameters, $\alpha$ and $\beta$, regulate the weighting of references in the reconstruction process; decreasing $\alpha$ and increasing $\beta$ result in a more stylized output. These adjustments afford \frameworkname{} enhanced adaptability across a diverse array of usage scenarios.
    \subsection{Why $\gamma = -\infty$}
    We set $\gamma = -\infty$ to simplify content and style disentanglement. As shown in the table Tab.\ref{tab:sen_abla_1}, using other values like 0.9, 1, or 1.1 lower due to the introduction of additional content-related features. With $\gamma = -\infty$, content and style are directly controlled by $\alpha$ and $\beta$, as supported by Figs.4(a), (b) and "+SDM" in Tab.2, where reconstruction difference are negligible.
    
    \subsection{Effect of $(\alpha, \beta)$}
\begin{figure}
\centering
\includegraphics[width=\linewidth]{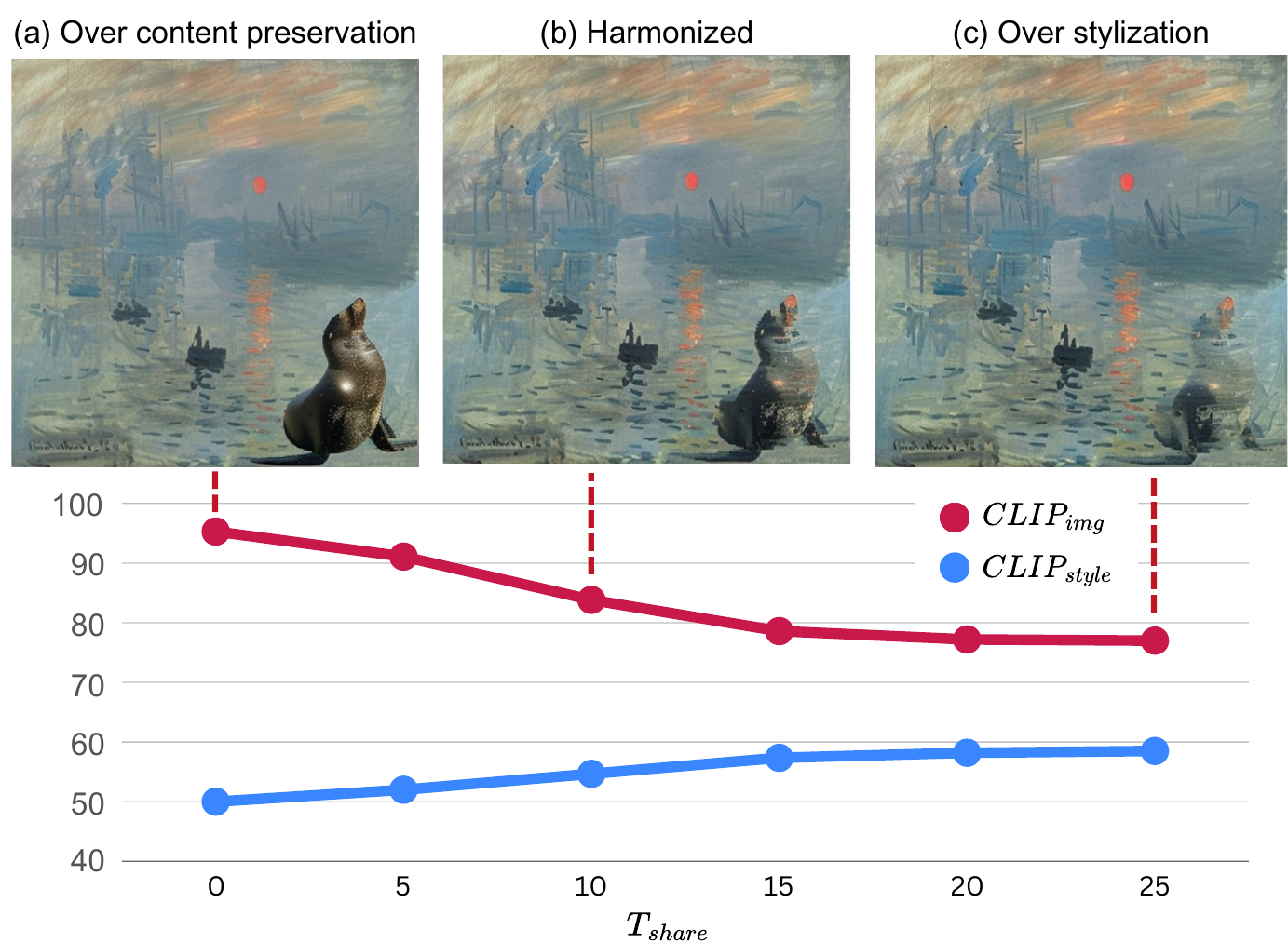}
\caption{Comparison of different stylized strength, when adjusting $T_{share}$ only.}
\label{fig:metrics_demo}
\end{figure}

\begin{figure}
\centering
\includegraphics[width=\linewidth]{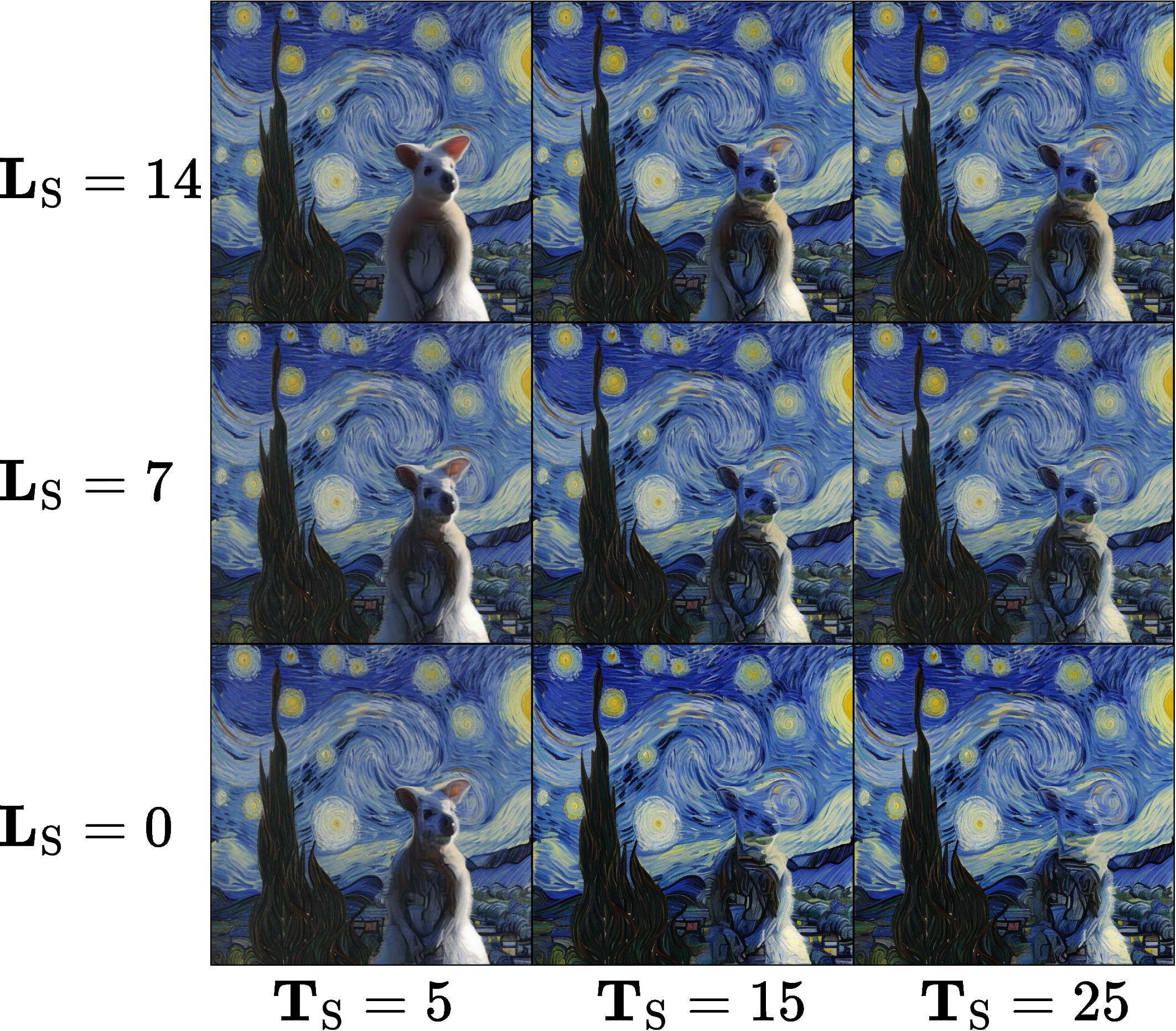}
\caption{Comparing various levels of stylized strength by adjusting both $T_{\text{share}}$ and $L_{\text{share}}$ (abbreviated as "S"), with fixed values for $\alpha=0.9$ and $\beta=1.1$.}
\label{fig:LS_demo}
\end{figure}

\begin{figure}
\centering
\includegraphics[width=\linewidth]{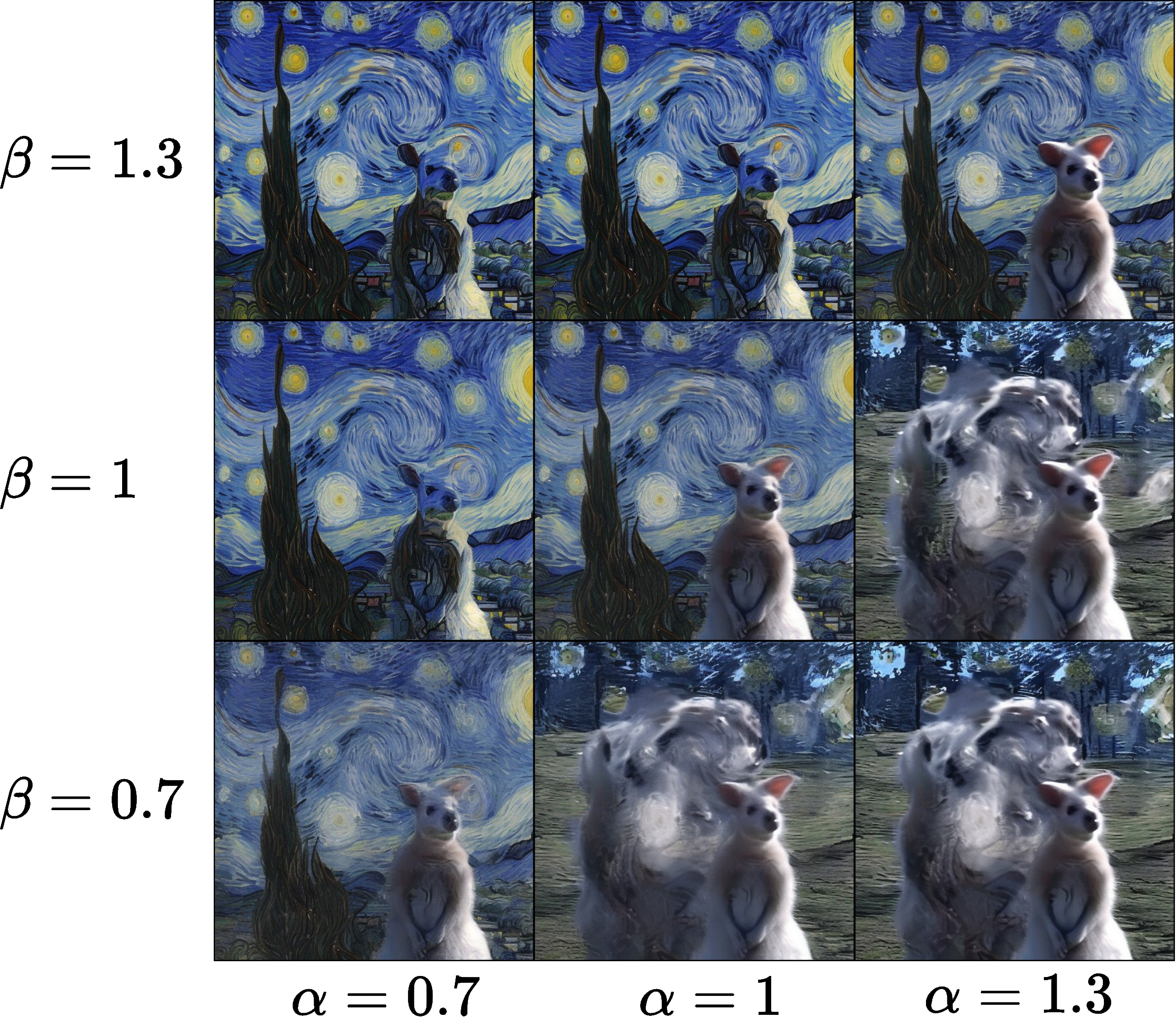}
\caption{Comparing various levels of stylized strength by adjusting both $\alpha$ and $\beta$, with fixed values for $T_\text{share}=15$ and $L_\text{share}=7$.}
\label{fig:ab_demo}
\end{figure}

    The quantitative results are shown Tab.\ref{tab:sen_abla_2}, with qualitative examples shown in \Figref{fig:ab_demo} and our project page. We observe that decreasing $\alpha$ or $\beta$ leads to grayish tones, while increasing them results in over-saturated colors with minimal stylization gains. This is reflected quantitatively: other settings disrupt background consistency, increasing with minor improvement. Thus, we chose $(\alpha, \beta) = (0.9, 1.1)$ to maintain background consistency while enhancing object stylization, which are the goal of painterly harmonization. Also we present a straightforward visualization depicting the difference by varying $T_{share}$ as shown in \Figref{fig:ab_demo} $T_\text{share}$ in \Figref{fig:metrics_demo}, alongside the qualitative outcome of altering others \Figref{fig:LS_demo} and .
    
    \begin{table}\
      \centering
      \begin{tabular}{lccc}
        \toprule
          $(\alpha, \beta, \gamma)$& $LP_{bg}$ & $LP_{fg}$ & $CP_{style}$\\
        \midrule
        $(1,1,-\infty)$ & 0.11 & 0.10 & 47.50\\
        $(0.9, 1.1,-\infty)$ & 0.12 & 0.32 & 56.37\\
        \midrule
        $(1,1,1)$ & 0.11 & 0.10 & 47.50\\
        $(0.9,1.1,0.9)$ & 0.12 & 0.25 & 52.33\\
        $(0.9,1.1,1)$ & 0.11 & 0.13 & 48.71\\
        $(0.9,1.1,1.1)$ & 0.11 & 0.10 & 47.61\\
        \bottomrule
      \end{tabular}
      \caption{Ablation study on the additional entry $\gamma$. The results indicate that incorporating $\gamma$ does not further expand the lower or upper bounds.}
      \label{tab:sen_abla_1}
    \end{table}
    
    \begin{table}\
      \centering
      \tiny
      \begin{tabular}{lcc|cccc}
        \toprule
          $(\alpha, \beta)$ &  (1,1) & (0.9, 1.1) & (0.9,1.5) & (0.5,1.5) & (1.5, 1.5) &  (0.5, 0.5)\\
         \midrule
         $LP_{bg}$ & 0.11 & 0.12 & 0.18 & 0.19 & 0.22 & 0.26\\
         $LP_{fg}$ & 0.10 & 0.32 & 0.36 & 0.37 & 0.14 & 0.27\\
         $CP_{style}$ & 47.50 & 56.37 & 55.96 & 56.48 & 45.64 & 49.51\\
         \bottomrule
       \end{tabular}
       \caption{Ablation study on different settings of stylization strength. We observed that increasing the difference between $(\alpha, \beta)$ marginally enhances stylization strength but introduces significantly more noise, compromising background preservation.}
       \label{tab:sen_abla_2}
    \end{table}
     
\section{Appendix: More Applications}\label{app:more_applications}
These examples emerged as part of our broader exploration into the full potential of our proposed TF-GPH method. As our design has the property of harmoniously mixing the features from two different images by scaling the similarity (rather than directly adjusting attention output), we can perform tasks such as semantic mixing and exemplar-based inpainting. These toy examples are included in the Appendix to share results that might inspire similar lines of research and also as part of our future work.
 \subsection{Inpainting}
\begin{figure}[h]
\centering
\includegraphics[width=\linewidth]{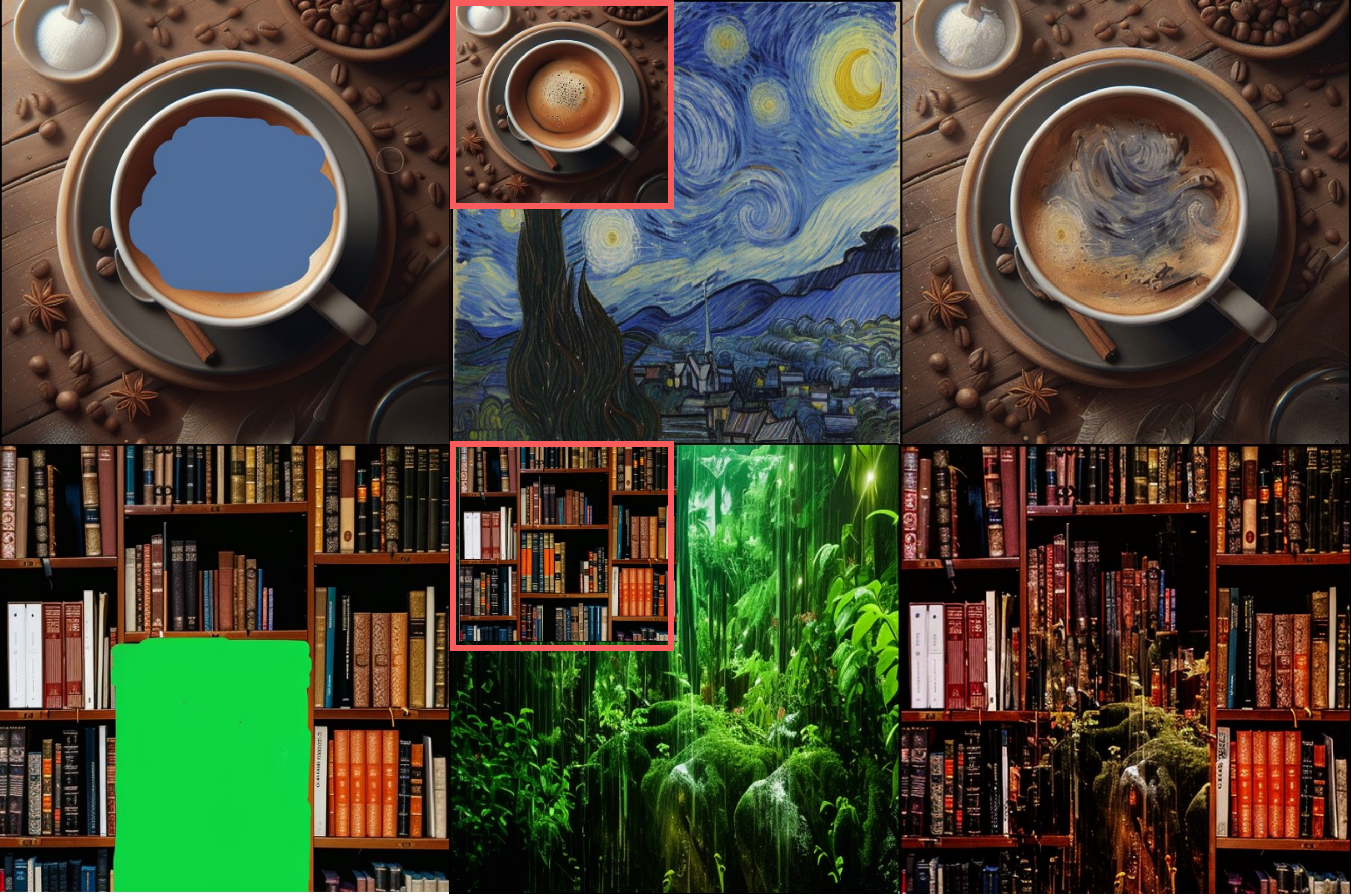}
\caption{The left column is the given $I^\text{comp}$ and corresponding inpaint mask, the middle column is the additionally provided $I^\text{f}$ (in \textcolor{red}{red box}) and $I^\text{b}$. The last column is the output image.}
\label{fig:inpaint_demo}
\end{figure}

 We incorporate our proposed share-attention w/ reweighting into the inpainting method Repaint\cite{repaint}. And provide the functionality of exemplar guided inpainting. We replaced the noisy latent inside mask area of $z^\text{comp}_T$ to the noise of $z^\text{b}_T$. The result can be found in \Figref{fig:inpaint_demo}, showing that our proposed method is able to provide current inpaint method additional exemplar information based on existing framework.
 
 \subsection{Semantic mixing}
\begin{figure}[h]
\centering
\includegraphics[width=\linewidth]{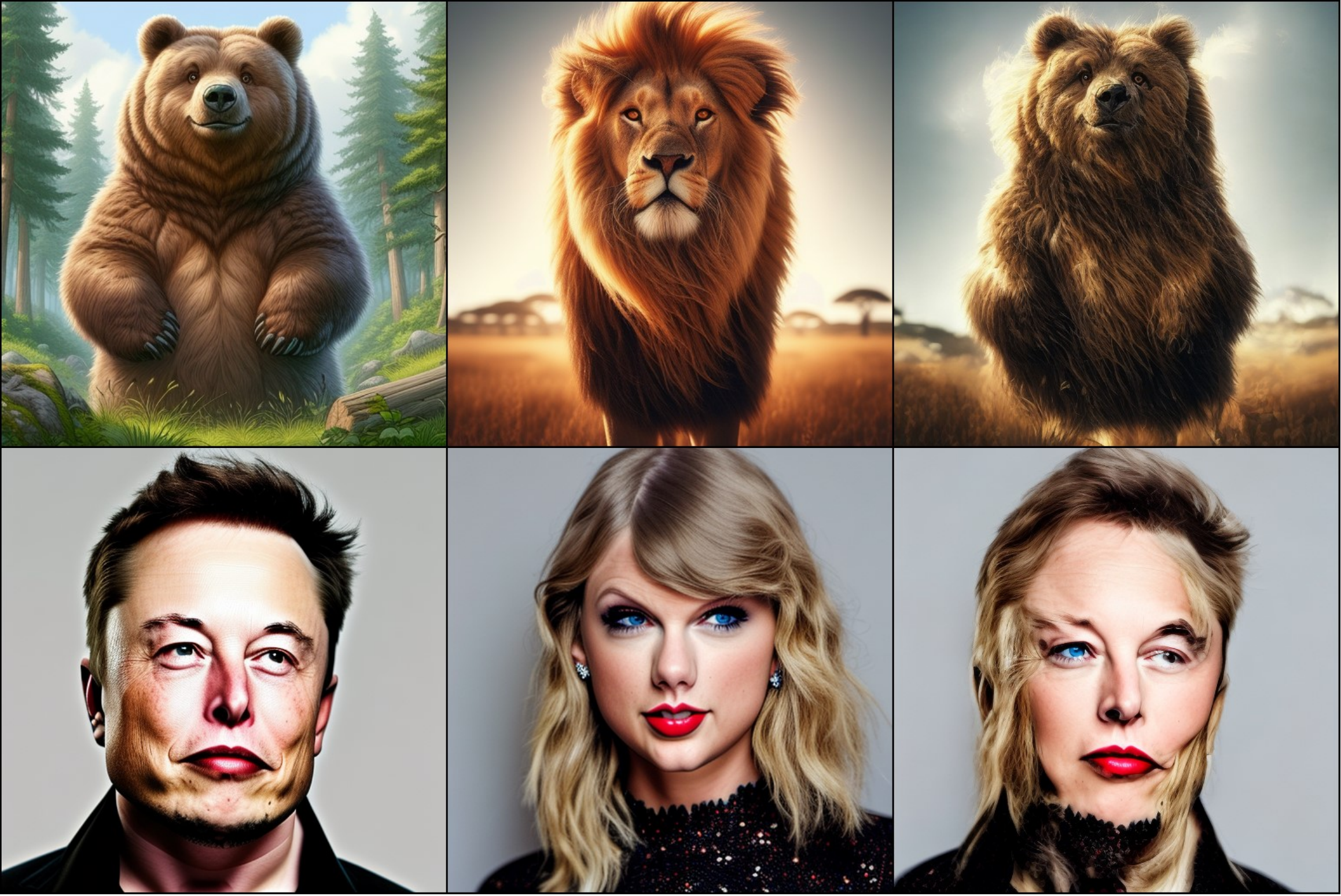}
\caption{The columns, from left to right, represent $I^\text{f}$, $I^\text{b}$, and $I^\text{o}$, where our share-attention layer is able to perform astonishing semantic mixing when combined with corresponding method.}
\label{fig:mix_demo}
\end{figure}

\begin{figure*}
  \centering
  \includegraphics[width=\linewidth]{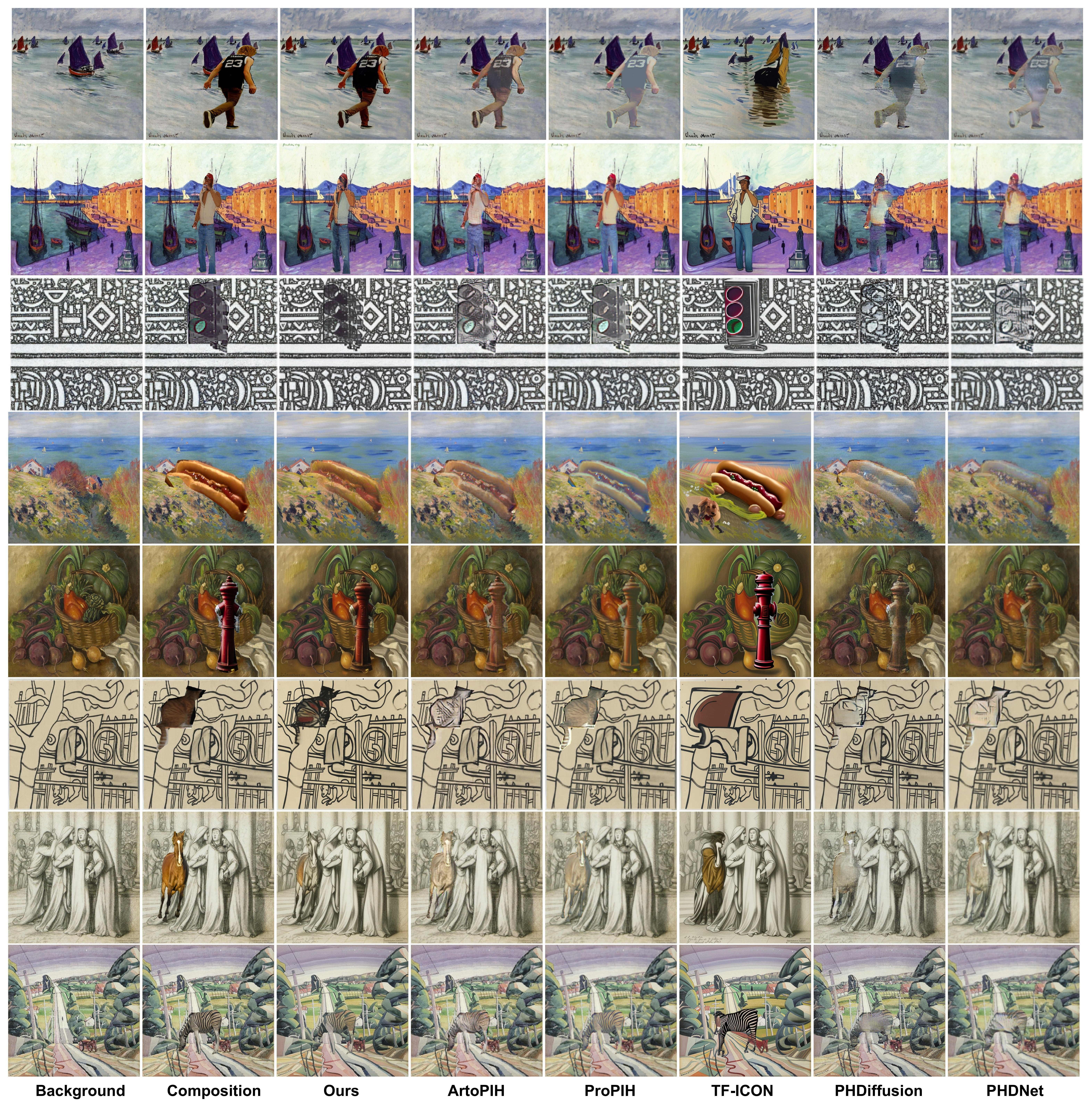}
  \caption{Qualitative comparison of WikiArt combined with COCO} 
  \label{fig:comparison_wikiart}
\end{figure*}

\begin{figure*}
  \centering
  \includegraphics[width=\linewidth]{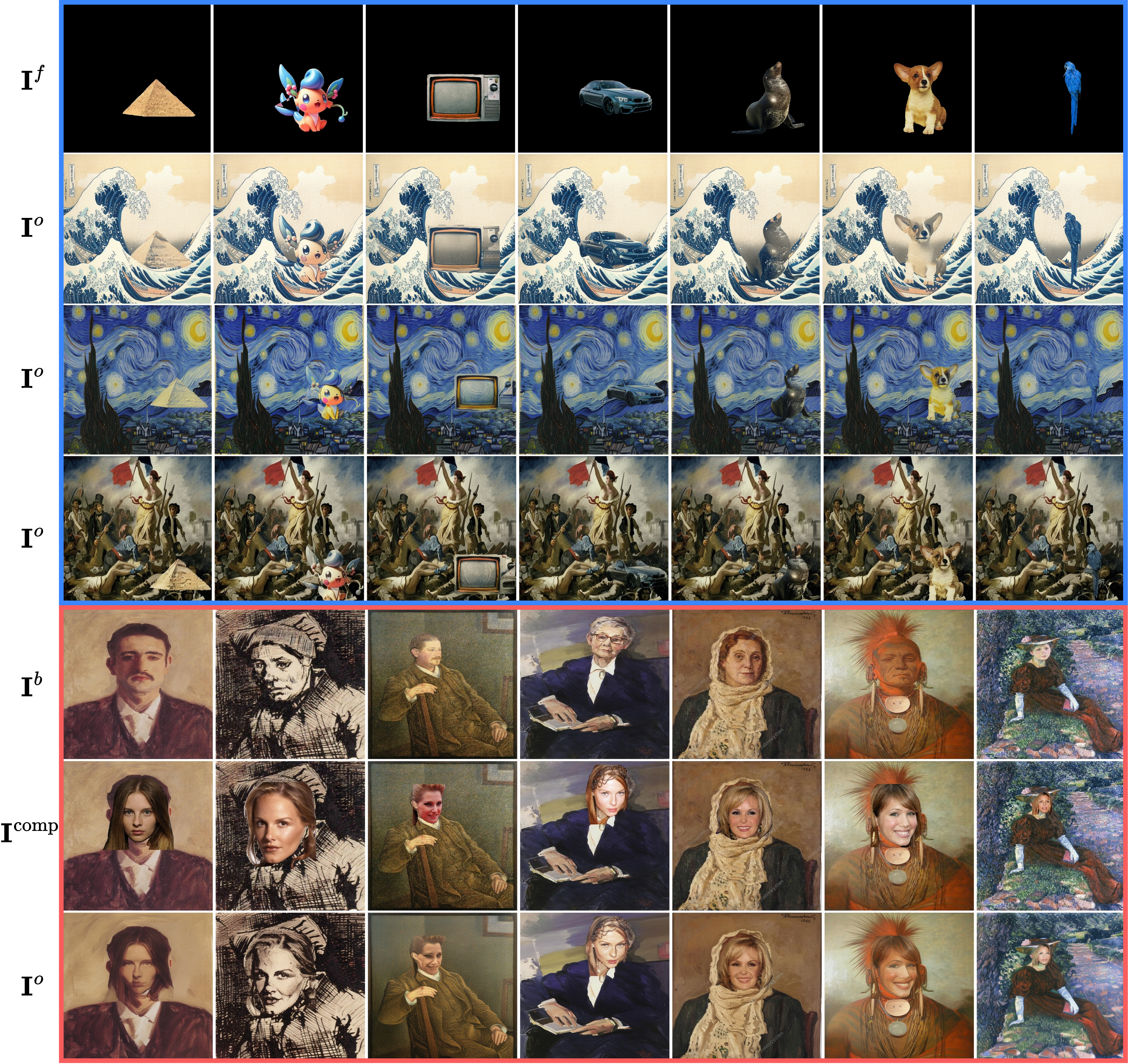}
  \caption{More painterly harmonization result of our proposed \frameworkname{} on GPH Benchmark. \textcolor{blue}{The row 1, is the input foreground objects ,and the row 2,3,4 are the corresponding outputs}, \textcolor{red}{The row 5 is the input background objects , the row 6 is the composite image with given face to paste, and the row 7 is the corresponding outputs}}
  \label{fig:more_gph_mix}
\end{figure*}

\begin{figure*}[h]
  \centering
  \includegraphics[width=\linewidth]{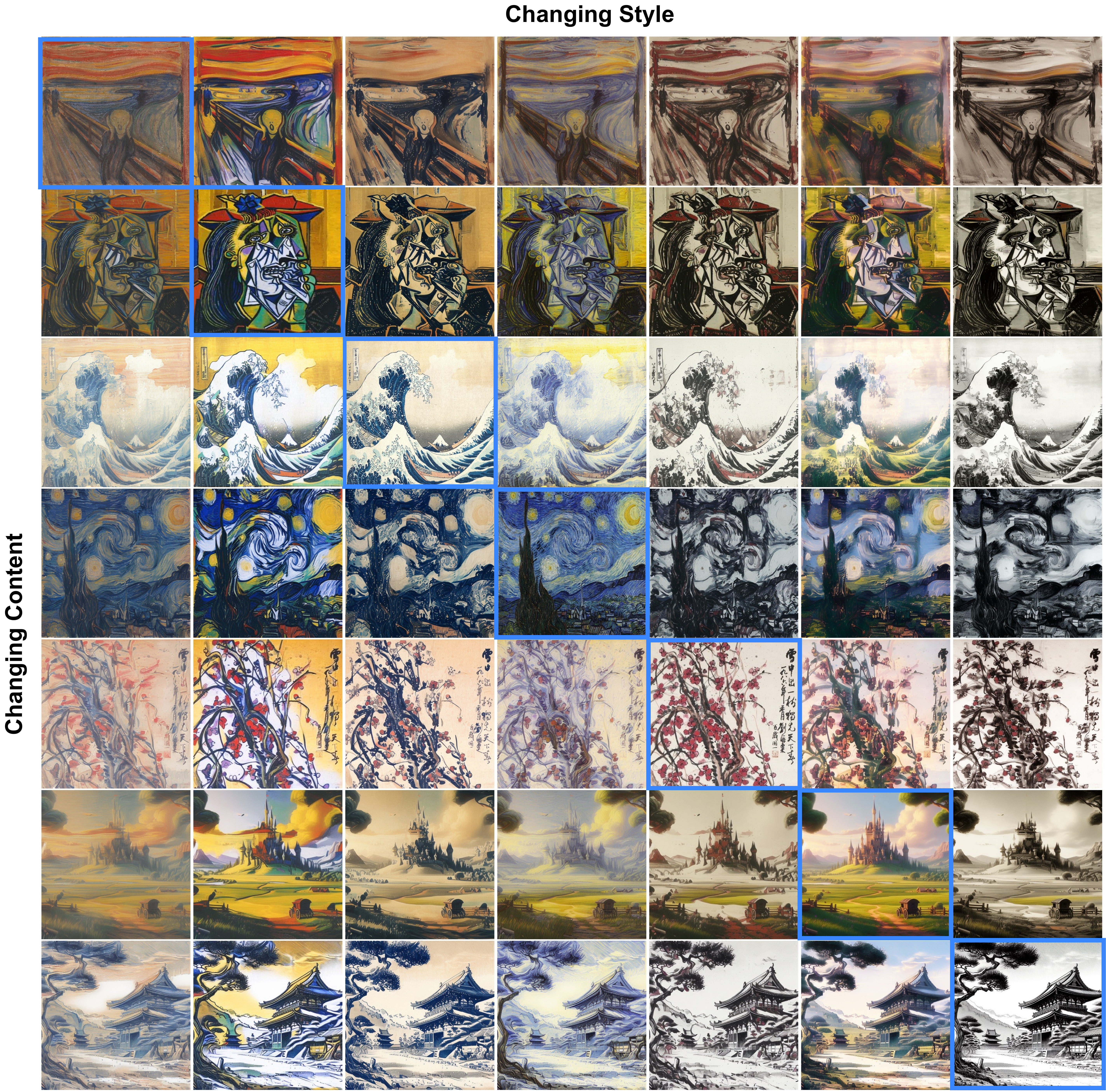}
  \caption{More style transfer result of our proposed \frameworkname{} on GPH Benchmark.  \textcolor{blue}{The images inside blue box, serves as both the \textbf{content reference for the corresponding ROW} and the \textbf{style reference for the corresponding COLUMN}},}
  \label{fig:more_gph_style}
\end{figure*}

 We test the compatibility of share-attention layer with semantic mixing method InjectFusion \cite{content_injection}, which is the adaption of the renowned Asyrp\cite{diffusion_semantic} approach. The core concept shared by these methods involves blending the semantic information from two images by manipulating their h-space, specifically the intermediary attention layer within the diffusion UNet architecture. To integrate the shared-attention layer into the InjectFusion, we simply allow the output image to attend to additionally provided $I^\text{b}$ via our share-attention layer. The generated result can be found in \Figref{fig:mix_demo}

\begin{table*}[t]
  \centering
  \small
  \caption{Quantitative results of GPH-Benchmark (\textsuperscript{\dag} represents the method with inference-time-adjustable hyperparameters. The left side of / represents content emphasis strategy, while the right side of / represents stylized emphasis strategy.)}
  \label{tab:main_gph}
  \begin{tabular}{c|ccccc|cccc} 
    \hline
     & \multicolumn{5}{c|}{Painterly Harmonization (256x256)} & \multicolumn{4}{c}{Style Transfer (256x256)} \\
     \hline 
    & Ours\textsuperscript{\dag} & ArtoPIH & ProPIH\textsuperscript{\dag} & TF-ICON\textsuperscript{\dag} & PHDiff\textsuperscript{\dag} & Ours\textsuperscript{\dag} & StyleID\textsuperscript{\dag} & Z-STAR\textsuperscript{\dag} & StyTr$^2$ \\
    \hline
    $Venue$  & - & AAAI'24 & AAAI'24  & ICCV'23  & MM'23  & - & CVPR'24 & CVPR'24 & CVPR'22 \\
    $LP_\text{bg} \downarrow$  & 0.07/0.07 & \textbf{0.05} & 0.06/0.06 & 0.16/0.32 & 0.06/0.06 & 0.66/\textbf{0.47} & 0.55/0.56 & 0.63/0.63 & 0.55  \\
    $LP_\text{fg} \downarrow$  & \textbf{0.04}/0.26 & 0.22 & 0.18/0.30 & 0.29/0.33 & 0.06/0.31 & \textbf{0.05}/0.40 & 0.29/0.42 & 0.12/0.33 & 0.33 \\
    $CP_\text{img} \uparrow$ & \textbf{97.58}/84.41 & 90.58 & 93.30/83.81 & 90.26/87.85  & 97.46/73.65 & \textbf{98.06}/75.55 & 84.65/71.64 & 93.33/71.18 & 84.32 \\
    $CP_\text{style} \uparrow$  & 48.74/\textbf{54.65} & 50.82 & 49.53/52.43 & 48.46/48.48 & 49.04/53.96 & 61.38/75.52 & 69.22/\textbf{78.96} & 59.61/69.50 & 64.98\\
    $CP_\text{dir} \uparrow$ & 0.01/9.11 & 3.69 & 2.03/10.09 & 2.66/3.91   & 0.35/\textbf{12.39}  & 2.08/45.02 & 24.92/\textbf{47.68} & 8.44/32.84 & 20.42 \\
    \hline

    \bottomrule 
  \end{tabular}
\end{table*}

\begin{table*}
  \centering
  \small
  \caption{Quantitative results of WikiArt w/ COCO  (\textsuperscript{\dag} represents the method with inference-time-adjustable hyperparameters. The left side of / represents content emphasis strategy, while the right side of / represents stylized emphasis strategy.)}
  \label{tab:main_coco512}
  \begin{tabular}{c|cccccccc} 
    \hline
     & \multicolumn{8}{c}{Painterly Harmonization (512x512)} \\
     \hline 
    & Ours\textsuperscript{\dag} & ArtoPIH & ProPIH\textsuperscript{\dag} & TF-ICON\textsuperscript{\dag} & PHDiff\textsuperscript{\dag} & PHDNet & SDEdit & DIB\\
    \hline
    $Venue$  & - & AAAI'24 & AAAI'24  & ICCV'23  & MM'23 & AAAI'23 & ICLR'22 & WACV'20\\  
    $LP_\text{bg} \downarrow$  & \textbf{0.08}/0.10 & 0.24 & 0.29/0.29 & 0.21/0.34 & 0.08/0.11  & 0.34 & 0.36 & 0.11 \\
    $LP_\text{fg} \downarrow$  & \textbf{0.10}/0.27 & 0.30 & 0.25/0.37 & 0.11/0.30  & 0.10/0.39 & 0.32 & 0.29 & 0.23 \\
    $CP_\text{img} \uparrow$ & \textbf{92.88}/80.40 & 83.55 & 86.94/78.33 & 83.79/82.65  & 91.78/80.32& 81.65 & 80.70 &  88.4  \\
    $CP_\text{style} \uparrow$  & 47.96/\textbf{55.25}  & 49.92 & 47.74/51.07 & 49.51/50.02 & 46.17/53.95 & 50.64 &  50.46& 48.59 \\
    $CP_\text{dir} \uparrow$ & 5.44/\textbf{19.20} & 13.04 & 9.20/18.11 & 7.63/11.18  & 6.49/18.14  & 15.80 &  13.46 & 8.67 \\
    \hline

    \bottomrule 
  \end{tabular}
\end{table*}

\begin{table*}
  \centering
  \small
  \caption{Quantitative results of WikiArt w/ COCO  (\textsuperscript{\dag} represents the method with inference-time-adjustable hyperparameters. The left side of / represents content emphasis strategy, while the right side of / represents stylized emphasis strategy.)}
  \label{tab:main_coco256}
  \begin{tabular}{c|cccccccc} 
    \hline
     & \multicolumn{8}{c}{Painterly Harmonization (256x256)} \\
     \hline 
    & Ours\textsuperscript{\dag} & ArtoPIH & ProPIH\textsuperscript{\dag} & TF-ICON\textsuperscript{\dag} & PHDiff\textsuperscript{\dag} & PHDNet & SDEdit & DIB\\
    \hline
    $Venue$  & - & AAAI'24 & AAAI'24  & ICCV'23  & MM'23 & AAAI'23 & ICLR'22 & WACV'20\\  
    $LP_\text{bg} \downarrow$  & 0.05/0.06 & \textbf{0.04} & 0.05/0.05 & 0.18/0.31 & 0.06/0.06  & 0.12 & 0.14 & 0.8 \\
    $LP_\text{fg} \downarrow$  & \textbf{0.06}/0.23 & 0.23 & 0.19/0.34 & 0.11/0.30  & 0.06/0.28 & 0.26 & 0.21 & 0.18 \\
    $CP_\text{img} \uparrow$ & \textbf{95.83}/83.31 & 87.31 & 90.84/81.62 & 83.79/82.65  & 94.28/78.66& 79.64 & 81.33 &  87.63  \\
    $CP_\text{style} \uparrow$  & 48.26/\textbf{54.41}  & 51.20 & 49.07/52.38 & 49.51/50.02 & 46.17/53.95 & 52.13 &  51.71& 51.43 \\
    $CP_\text{dir} \uparrow$ & 2.92/13.11 & 10.23 & 6.73/12.91 & 7.63/11.18  & 7.49/\textbf{21.11}  & 16.77 &  14.52 & 9.11 \\
    \hline

    \bottomrule 
  \end{tabular}
\end{table*}

\end{document}